\documentclass[conference]{IEEEtran}

\makeatletter
\def\ps@IEEEtitlepagestyle{%
  \def\@oddfoot{\mycopyrightnotice}%
  \def\@evenfoot{}%
}

\usepackage{blindtext}
\usepackage{eso-pic}
\IEEEoverridecommandlockouts
\usepackage{cite}
\usepackage{amsmath,amssymb,amsfonts}
\usepackage{algorithmic}
\usepackage{graphicx}
\usepackage{textcomp}
\usepackage{xcolor}

\usepackage{hyperref}
\usepackage{booktabs}

\def\BibTeX{{\rm B\kern-.05em{\sc i\kern-.025em b}\kern-.08em
    T\kern-.1667em\lower.7ex\hbox{E}\kern-.125emX}}
    
\usepackage{eso-pic}

\usepackage{fancyhdr}

\newcommand{\arxivheader}{%
  \small Accepted for publication in the 2026 International Conference on Electrical and Computer Engineering Researches (ICECER 2026)
}

\newcommand{\arxivcopyright}{%
  \vbox{%
    \hrule height 0.4pt \vskip 5pt
    \small \noindent © 2026 IEEE. Personal use of this material is permitted. Permission from IEEE must be obtained for all other uses, in any current or future media, including reprinting/republishing this material for advertising or promotional purposes, creating new collective works, for resale or redistribution to servers or lists, or reuse of any copyrighted component of this work in other works.
  }%
}

\fancypagestyle{arxivstyle}{
  \fancyhf{}
  
  \fancyhead[C]{\arxivheader}   
  \fancyfoot[C]{\arxivcopyright} 
}

\begin{document}
\title{\vspace*{1cm} Deep Learning-Based Road Lane Quality Assessment from Egocentric Dashcam Video
}

\author{
\IEEEauthorblockN{
Shaif Chowdhury,
Akhila Yeruvasu,
Moshiur Rahman Tonmoy,
Ayush Goyal,
and Dunren Che
}
\IEEEauthorblockA{
\textit{Department of Electrical Engineering and Computer Science, Texas A\&M University--Kingsville}\\
Kingsville, TX 78363, USA \\
\{shaif.chowdhury, ayush.goyal, dunren.che\}@tamuk.edu\\
\{akhila.yeruvasu, moshiur\_rahman.tonmoy\}@students.tamuk.edu
}
}

\maketitle
\thispagestyle{arxivstyle}

\begin{abstract}
Reliable lane markings are important for both human drivers and automated driving systems, but their condition is still commonly assessed through manual inspection. Existing autonomous-driving datasets primarily support lane detection and scene understanding and usually lack maintenance-oriented labels for lane-marking visibility, continuity, or degradation. 

In this study, we present a geo-referenced smartphone dashcam dataset and baseline framework for lane-marking quality assessment. The dataset contains 63,839 road-facing frames collected across South Texas using a GPS-enabled dashcam application. A subset of 3,339 frames is matched with OpenStreetMap-derived lane-count labels, which are used as weak supervision for lane-count regression and map-consistency evaluation. We propose three lane-quality measures: Contrast Stability Index (CSI), which captures lane-background contrast using Laplacian responses; Line Continuity Ratio (LCR), which estimates structural continuity using edge and Hough-line evidence; and Regression Lane Confidence Score (RLCS), which measures agreement between predicted and map-derived lane counts on matched frames. 

The baseline prediction models attained strong performance, with CSI (MAE $0.039$, $R^2=0.97$) and LCR (MAE $0.093$, $R^2=0.84$). Notably, CSI and LCR are no-reference, segmentation-derived quality metrics that can be computed without dense human labels. In a pilot validation using 50 human-ranked images, these proposed lane-specific metrics show stronger rank agreement with human lane-quality ordering (e.g., Spearman’s $\rho=0.7316$ and Kendall's $\tau=0.5472$ for LCR) than general image-quality measures such as PSNR, SSIM, BRISQUE, NIQE, and LPIPS. We believe the dataset, metrics, and baseline provide an initial step toward scalable, low-cost, location-aware lane-marking quality assessment. Data and code: \url{https://github.com/Shaif95/Road_Quality}
\end{abstract}


\section{Introduction}
\label{sec:intro}

\begin{figure}[t]
    \centering
    \includegraphics[width=\linewidth]{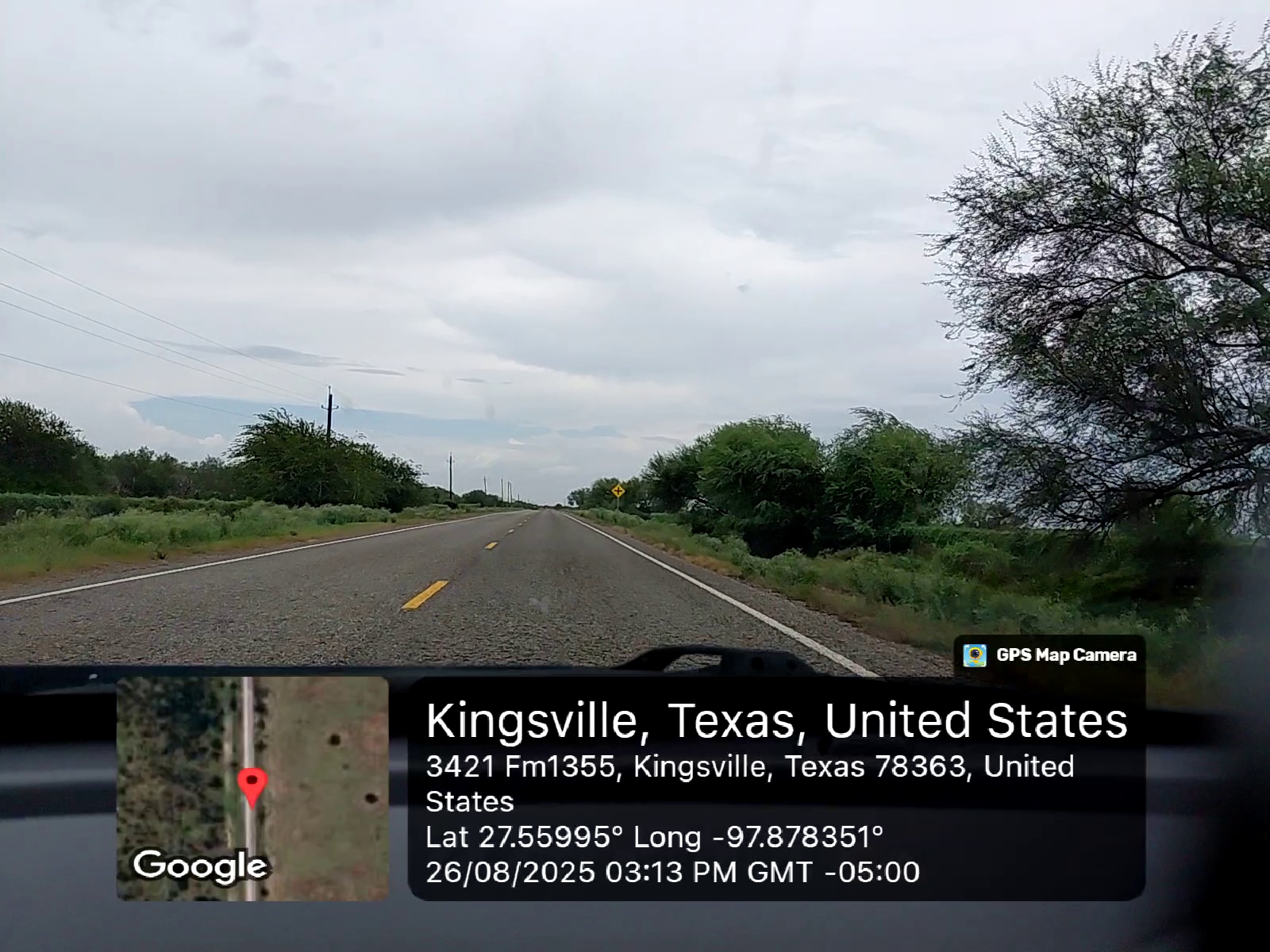}
\caption{
Example frame from our smartphone dashcam dataset, captured near Kingsville, Texas, USA. The image includes a metadata overlay generated by the GPS Map Camera application, which automatically embeds location-based information such as latitude (27.55995°), longitude (-97.87835°), and physical address (3421 FM1355, Kingsville, TX 78363, USA). It also records the capture time (26 August 2025, 03:13 PM GMT–05:00), providing spatiotemporal context for each frame. This overlay enables direct mapping between visual lane conditions and real-world geospatial coordinates, facilitating integration with external mapping APIs and supporting large-scale, location-aware evaluation of road and lane quality.}
    \label{fig:overlay}
\end{figure}

Lane markings\cite{tidjani2019impact} are important for providing guidance to human drivers, maintaining traffic flow, and have been identified as one of the most important infrastructures for autonomous vehicles\cite{marciano2024effect}. However, in general, lane markings often wear out over time due to traffic on the road, and that can lead to more road hazards. Due to the importance of lane markings, continuous monitoring and fixing are required to maintain minimum quality standards\cite{foucher2011detection}. But in general, monitoring of lane markings is often done by blocking traffic and leads to road closures, and in other cases, it also leads to traffic workers being put in dangerous situations. While a lot of existing research is based on autonomous vehicles\cite{kato2015open} ability to detect these lane lines\cite{behrendt2019unsupervised}, there is very little research on monitoring of lane lines for maintenance of quality standards\cite{pasetto2013experimental}. In this paper, we aim to address this by creating an open dataset that has road video as well as details of location, time, and other data that could be useful for assessing the quality of lane markings. We also develop deep learning-based models and metrics for the assessment of lane line quality. 

\subsection{Smartphone-Based Lane Quality Assessment}

Existing lane detection methods accurately localize lane markings but are not designed to evaluate their visibility, continuity, or overall quality. Commodity smartphones equipped with dashcam applications provide a scalable and low-cost alternative for collecting geo-tagged road videos during normal driving. Recent progress in deep learning\cite{he2016identity,kuutti2020survey,chowdhury2024efficient,liu2015very,dosovitskiy2020image,chowdhury2024active,arnab2021vivit,chowdhury2022recognition,ren2021applying,tonmoy2025leveraging,krizhevsky2009learning,chowdhury2025supervised,tonmoy2024lightweight, krizhevsky2012imagenet} and computer vision now makes it possible to extract meaningful information from such smartphone imagery, enabling the smartphone-based approach to lane quality assessment developed in this work. In this work, we use the \textit{GPS Map Camera} application to capture synchronized video and GPS metadata, enabling integration with OpenStreetMap to obtain lane-count information and develop automated metrics for lane quality assessment, including lane visibility, continuity, and semantic correctness.
\subsection{Overview of Our Work}

We present the \textbf{Smartphone Dashcam Road Quality dataset}, built from {23 video sequences} captured across South Texas using GPS-enabled smartphones. The data includes \textbf{63,839 valid geo-tagged frames}, allowing spatial analysis of lane quality. Using the \textbf{OpenStreetMap (OSM) API}, we retrieve lane count information based on GPS coordinates\cite{haklay2008openstreetmap}. Figure~\ref{fig:overlay} illustrates a representative example. We treat these OSM lane-count labels as weak map-derived supervision rather than direct visual lane-quality ground truth. Frames without lane data, typically in rural areas, are excluded. The resulting dataset pairs each frame with its corresponding lane count, supporting a correctness metric based on lane line prediction accuracy. {We also propose three quantitative metrics for the quality of the lane: }

\begin{enumerate}

    \item \textbf{Regression Lane Confidence Score (RLCS)}: The metric captures the semantic reliability of lane structure using a lane count confidence measure.
    
    \item \textbf{Contrast Stability Index (CSI)}: Quantifies lane visibility and contrast stability using Laplacian-based high-frequency responses.
    
    \item \textbf{Line Continuity Ratio (LCR)}: Measures geometric continuity of lane markings via Hough-based line aggregation.
    
\end{enumerate}

Our framework enables both \textbf{image-level} and \textbf{video-level} analysis, offering a scalable, data-driven foundation for evaluating lane visibility and continuity.

\subsection{Contributions}

Our main contributions are summarized as follows:

\begin{itemize}

\item \textbf{First open benchmark for lane quality assessment}: {To the best of our knowledge,} we introduce the first publicly available benchmark for lane marking quality assessment, consisting of a data set and evaluation framework that integrate egocentric smartphone dashcam videos with synchronized GPS data. The data set comprises more than 63,000 geo-referenced frames collected across roadways in South Texas.

\item \textbf{{Transparent} lane-quality metrics}: We introduce three complementary metrics for lane quality assessment: the \textit{Contrast Stability Index (CSI)}, which captures local lane–background contrast using Laplacian responses; the \textit{Line Continuity Ratio (LCR)}, which measures geometric continuity via Hough-based line aggregation; and the \textit{Regression Lane Confidence Score (RLCS)}, a regression-based semantic reliability measure that quantifies confidence in lane structure without discrete classification.

\item \textbf{Human-validated benchmark and evaluation}: We validate the proposed metrics using a {pilot} human-ranked benchmark of 50 lane images and compare them against image quality methods such as PSNR, SSIM, BRISQUE, NIQE, and LPIPS. The proposed \textit{LCR} achieves the highest agreement with human rankings ($\rho=0.7316$, $\tau=0.5472$). The baseline prediction models also achieve strong performance, with CSI (MAE 0.039, $R^2=0.97$) and LCR (MAE 0.093, $R^2=0.84$).

\end{itemize}

\begin{figure*}[t]
    \centering
    \includegraphics[width=0.33\textwidth]{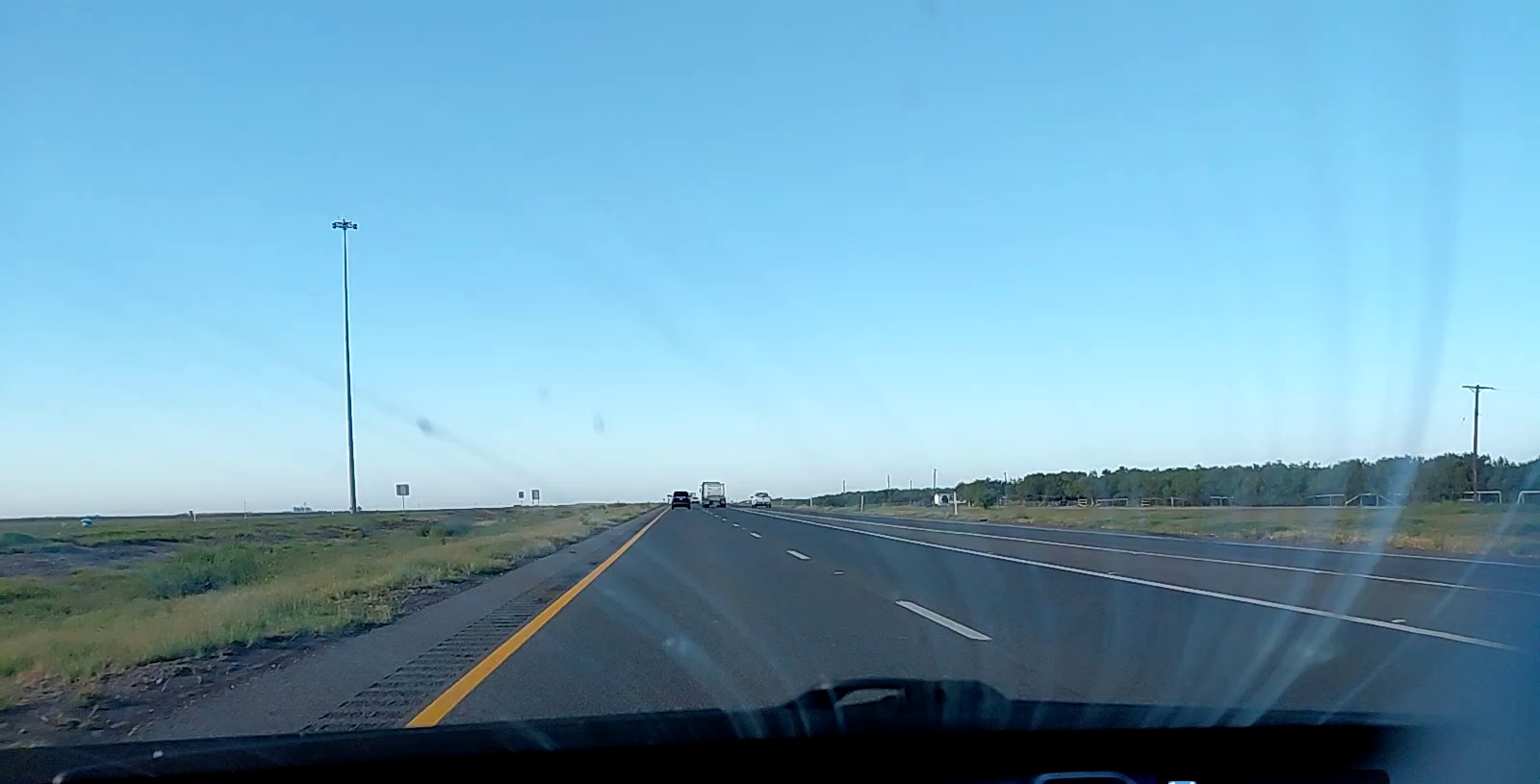}\hfill
    \includegraphics[width=0.33\textwidth]{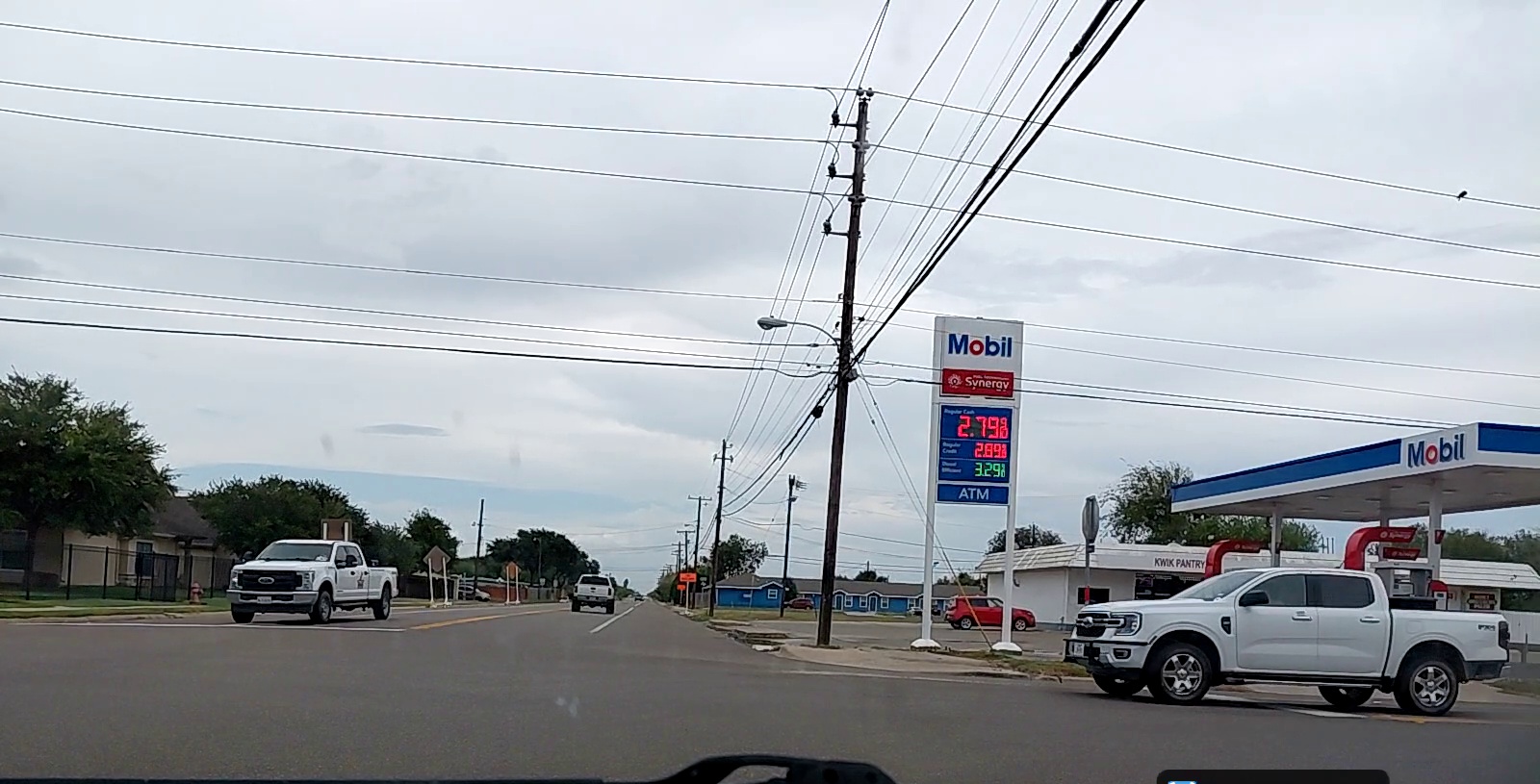}\hfill
    \includegraphics[width=0.33\textwidth]{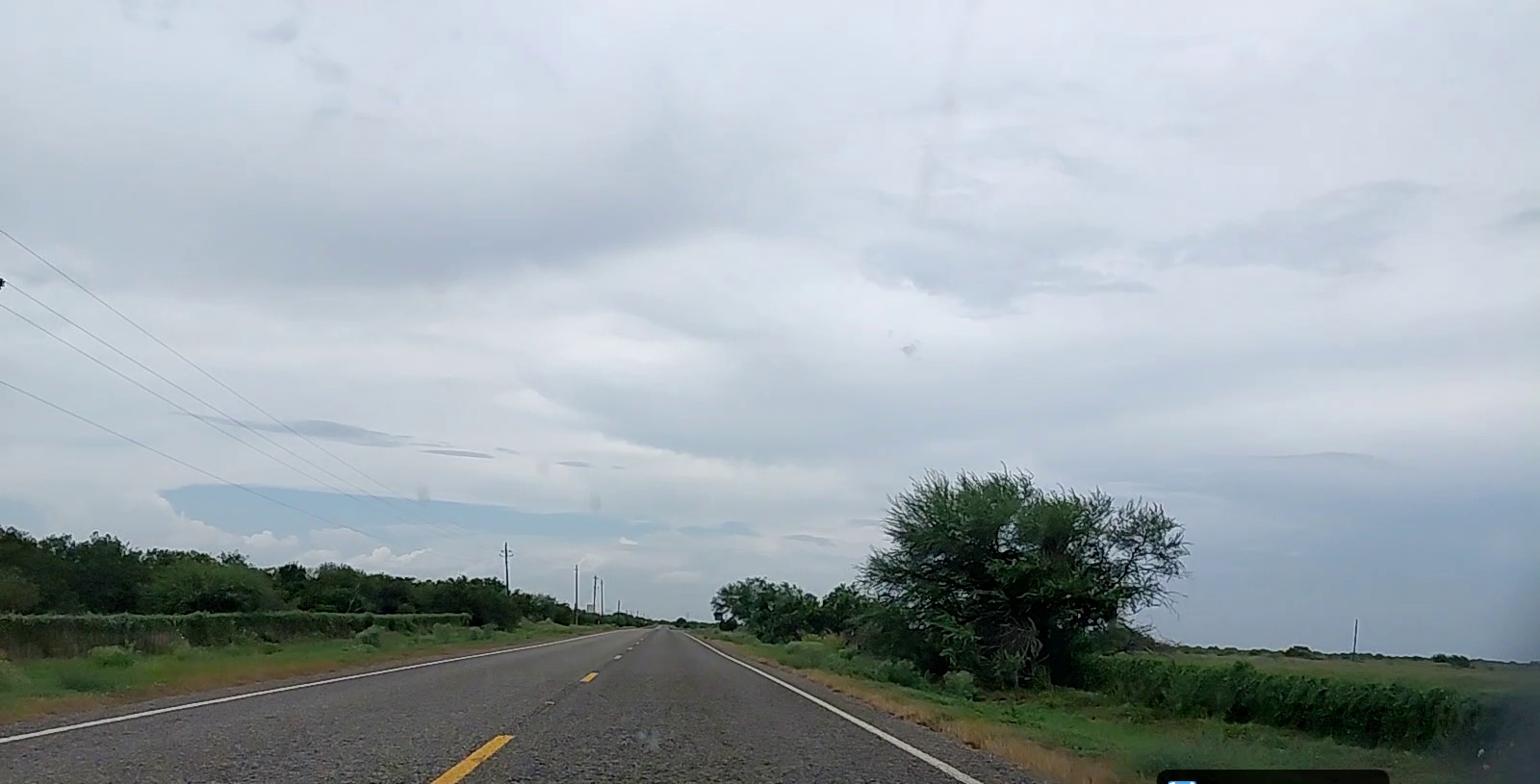}\\[1mm]
    \includegraphics[width=0.33\textwidth]{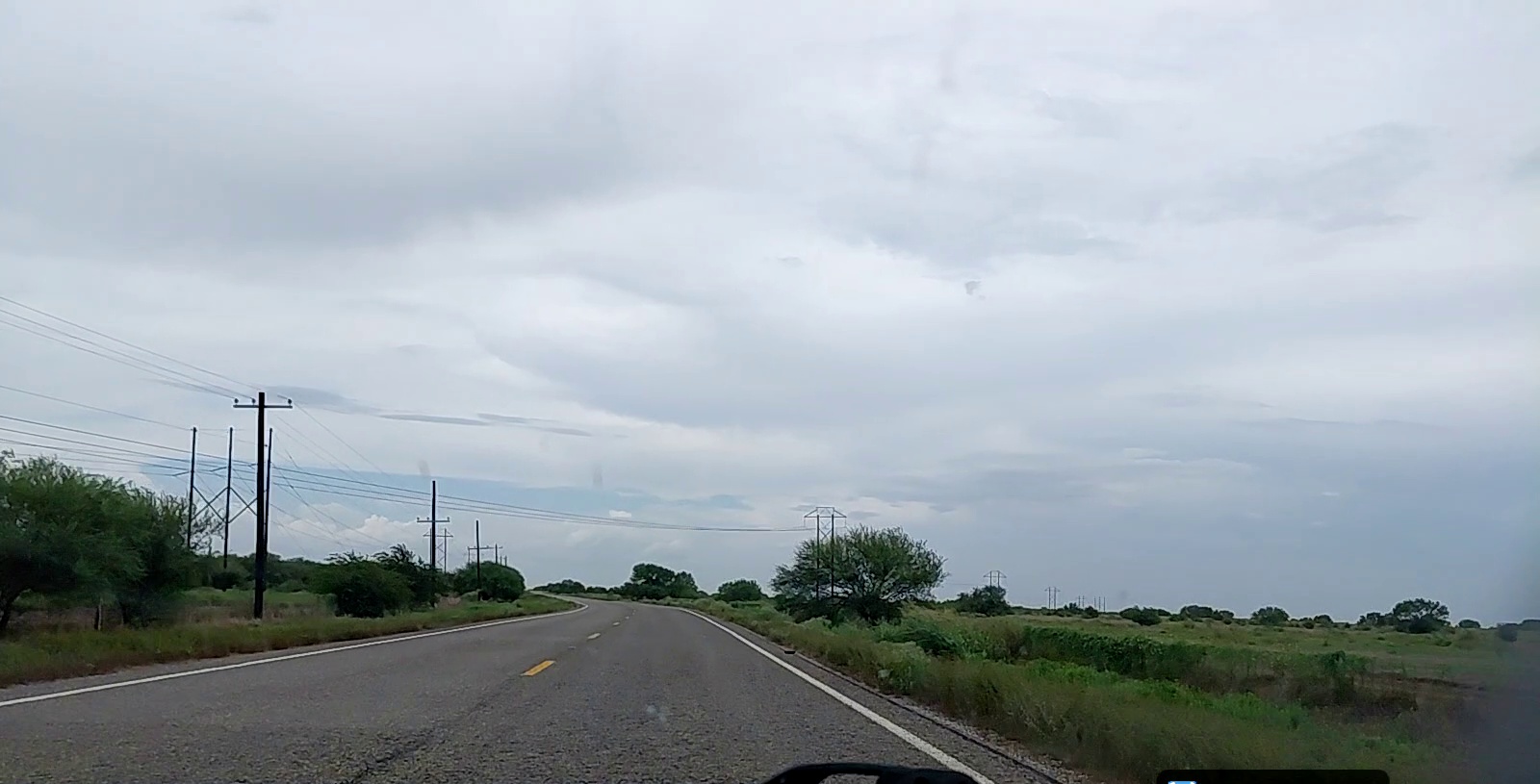}\hfill
    \includegraphics[width=0.33\textwidth]{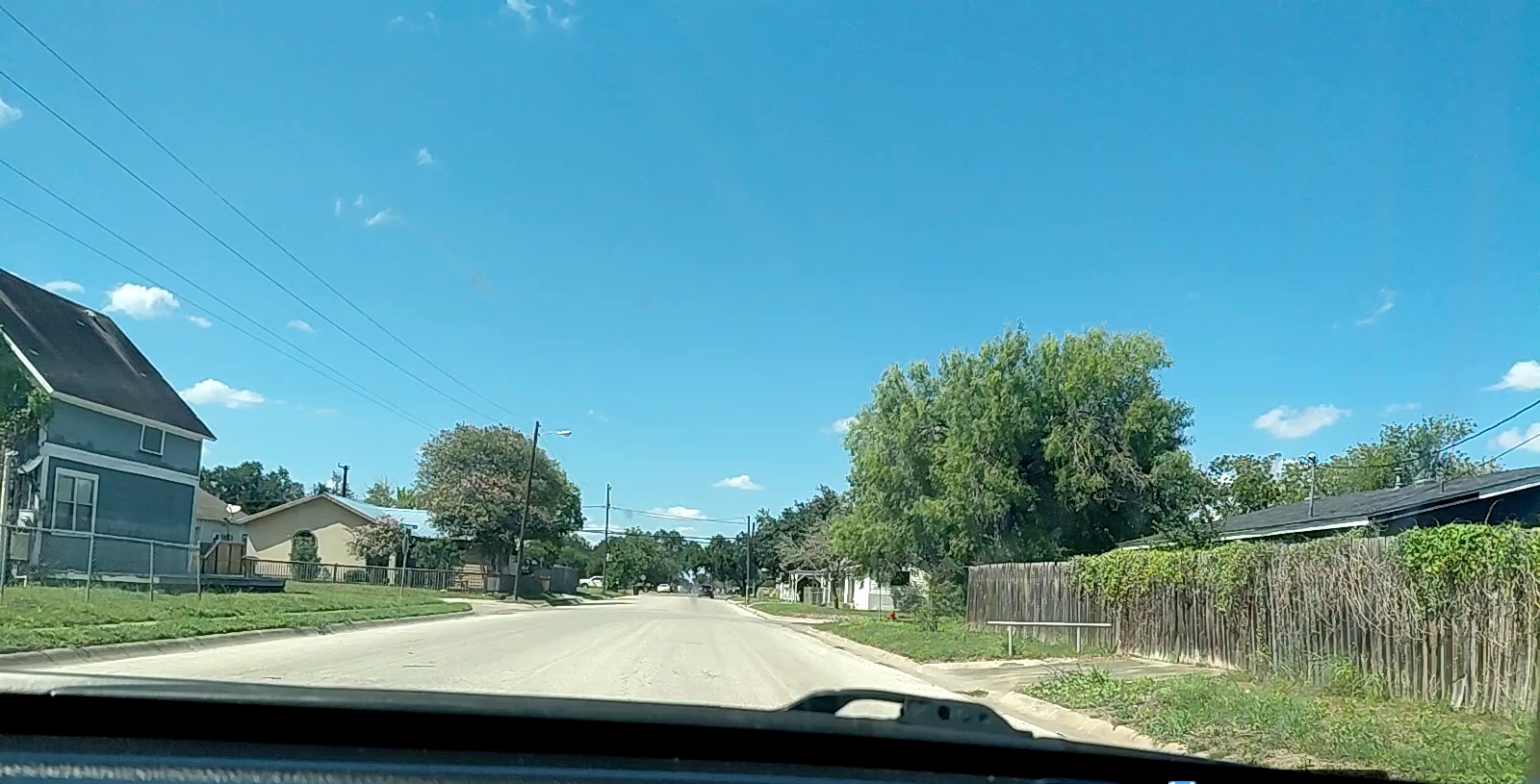}\hfill
    \includegraphics[width=0.33\textwidth]{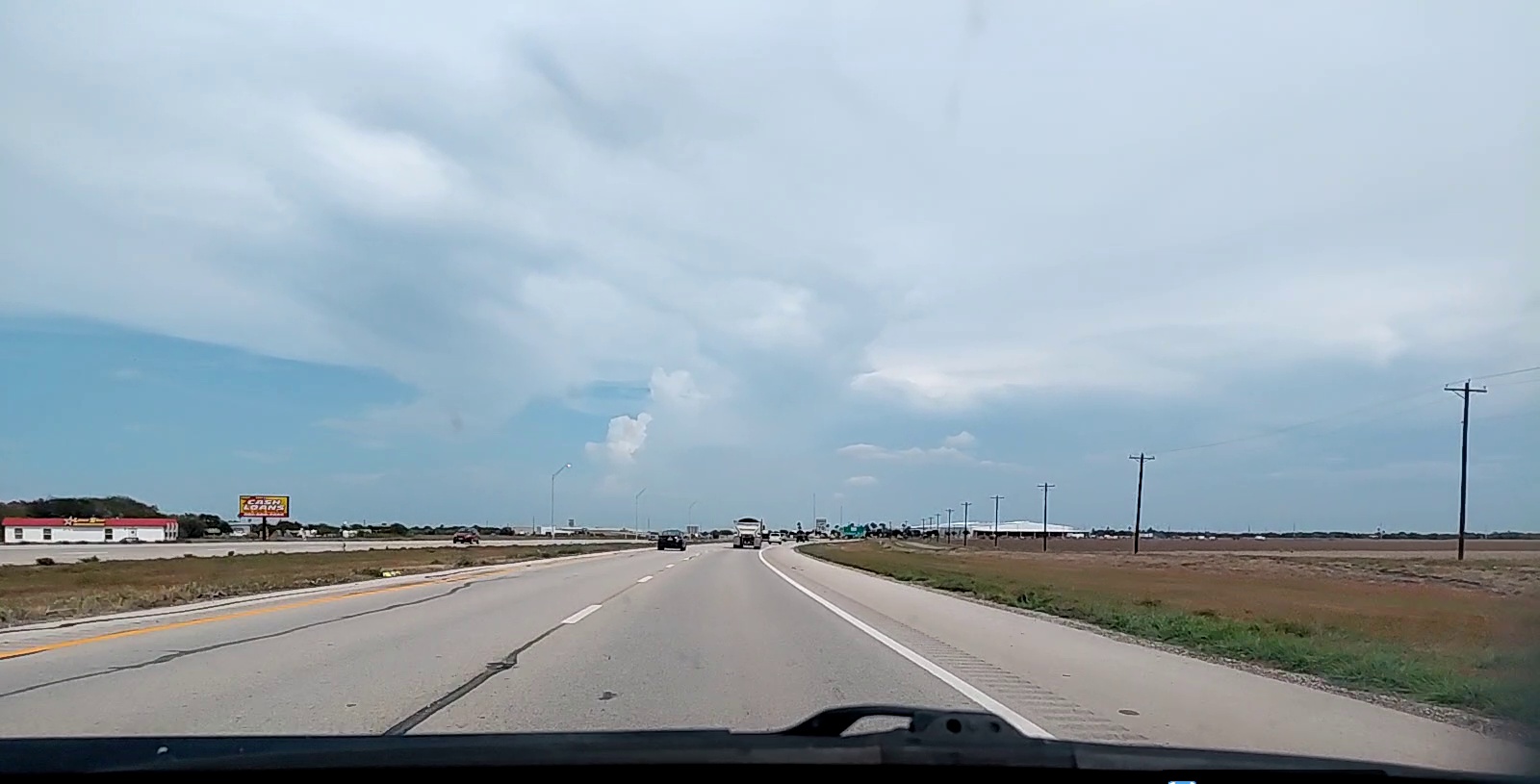}
    \caption{Representative smartphone dashcam scenes showing diverse roadway
    and lane-marking conditions used for lane quality assessment.}
    \label{fig:road_diversity}
\end{figure*}






\section{Related Work}
\label{sec:related}

Lane quality assessment is closely related to prior work in lane detection\cite{tang2021review}, roadway infrastructure monitoring\cite{masino2017learning}, transportation data sets, smartphone-based sensing, and image quality evaluation. While each research area has made significant progress, no existing approach provides a unified and scalable framework for measuring lane marking clarity, continuity, or degradation. This section summarizes the most relevant literature and highlights the gaps addressed by this study.

\subsection{Lane Detection and Segmentation}
Traditional lane detection methods are based on classical image-processing techniques such as the Canny edge detector~\cite{canny1986computational} and the Hough Transform~\cite{duda1972use}. Although effective for simple lane structures, these methods are sensitive to shadows, illumination changes, worn lane markings, and occlusions. Deep-learning-based approaches have significantly improved robustness by leveraging semantic representations. SCNN introduces spatial message passing to model long-range lane dependencies~\cite{pan2018spatial}, while LaneNet applies semantic segmentation and embedding-based clustering to identify individual lane instances~\cite{hou2019learning}. Large-scale data sets such as CULane have further enabled the development of reliable lane detection models in various driving environments~\cite{qin2020ultra}. However, despite achieving strong lane location performance, these methods do not quantify lane quality, including visibility, continuity, or the degree of degradation of lane markings.

\subsection{Road Surface and Infrastructure Quality Assessment}
Several studies have focused on using cameras, LiDAR, and smartphone sensors to assess general road surface condition. Vision-based approaches detect common pavement defects such as cracks, potholes, and surface irregularities. Abbas and Ismael demonstrated automated detection of cracks and potholes using image-processing techniques~\cite{abbas2021automated}, while Radopoulou and Brilakis developed a vision-based system capable of identifying multiple pavement distresses~\cite{radopoulou2017automated}. Smartphone-based systems use accelerometer and GPS data to detect anomalies; Eriksson et al.\ introduced a mobile sensing system for pothole detection~\cite{eriksson2008pothole}, and Sattar et al.\ reviewed several smartphone-based methods for identifying bumps and roughness~\cite{sattar2018smartphone}. LiDAR-based approaches have also been applied to measure pavement deformation, such as rutting~\cite{faisal2023rutting}. While effective for surface-level defects, these methods do not evaluate lane-marking visibility or continuity, indicating the absence of lane-specific quality assessment tools.

\subsection{Related Transportation Data Sets and Road Monitoring Approaches}

Existing transportation data sets, including KITTI and ApolloScape, have significantly advanced autonomous driving research by providing annotated road-scene imagery for tasks such as object detection, segmentation, and 3D scene understanding~\cite{geiger2012kitti,huang2018apolloscape}. However, these benchmarks are designed for perception tasks rather than for evaluating lane-marking visibility, continuity, or degradation, and they do not provide standardized annotations for lane quality assessment. Recent smartphone-based road monitoring has demonstrated that widely available mobile sensors enable scalable and cost-effective infrastructure assessment through crowdsourced data collection~\cite{eriksson2008pothole}. {Motivated by these, our work introduces a lane-marking video dataset by organizing smartphone dashcam video, GPS metadata, OSM-derived lane-count labels, and lane-specific quality metrics into a maintenance-oriented baseline framework.}

\section{Method}
\label{sec:method}

The objective of this paper is to create a reliable AI toolset for assessing the quality of road lane lines from vehicle dashcam footage. We use a mobile app to get GPS data along with video of the road. We evaluate all lane markings that are available within the view of the camera.

\subsection{Data Collection and Frame Extraction}
\label{sec:method:data}

We collect road-facing dashcam videos using a commodity Android smartphone (Samsung Galaxy S16 PM camera) running \textit{GPS Map Camera} (v1.7.2). The phone is mounted on the vehicle's windshield in a fixed, forward-facing orientation. We assume the camera intrinsics are pre-calibrated by the device and that non-linear lens distortions are minimal in our operating regime. We further assume that the GPS signal provided by the application is temporally synchronized with the captured video stream, enabling frame-level alignment between visual observations and geo-coordinates.

\paragraph{Notation.}
Common notations used throughout the paper are defined as follows:

\begin{itemize}
    \item \(V = \{I_t\}_{t=1}^{T}\), a video sequence of length \(T\).
    \item \(I_t \in \mathbb{R}^{H \times W \times 3}\), the RGB image frame at time \(t\).
    \item \(\tilde{\mathbf{x}} = [\mathbf{x}^\top, 1]^\top\), the homogeneous representation of a vector \(\mathbf{x}\).
    \item \(\mathbf{p}_{k,t} = [u, v]^\top \in \mathbb{R}^2\), the \(k\)-th pixel location in frame \(I_t\) (image coordinates).
    \item \(\mathbf{g}_t = [\phi_t, \lambda_t]^\top\), the GPS coordinate for frame \(I_t\), where \(\phi_t\) is latitude and \(\lambda_t\) is longitude.
\end{itemize}

\paragraph{Frame extraction.}
For each recorded video \(V\), we extract per-frame images \(\{I_t\}\) along with the corresponding GPS coordinates \(\{\mathbf{g}_t\}\) as provided by the application. Unless otherwise specified, each frame is treated as an independent sample for image-level metric computation. For learning-based baselines, frames are resized to a fixed spatial resolution (e.g., \(96 \times 96\)).

\paragraph {Dataset summary.}
We record \textbf{23 video sequences} across South Texas and extract frames with associated geospatial metadata. After filtering frames with unreliable or missing GPS, the dataset contains \textbf{63{,}839 valid frames}. The routes cover diverse environments, including urban multi-lane arterials and rural highways, spanning
\(27.0604^\circ\text{--}27.8830^\circ\) N and \(-97.8837^\circ\text{--}-97.8051^\circ\) W. Each frame is associated with latitude--longitude coordinates, enabling downstream mapping and spatial correlation of lane-quality metrics. The data set consists primarily of highway scenes (\textbf{60\%}), with the remaining frames captured on rural roads. All recordings were collected during daytime under clear or moderately cloudy conditions at typical driving speeds of \textbf{40--80 mph}. Figure~\ref{fig:road_diversity} presents representative road scenes from the dataset across diverse settings. 

\subsection{Geo-Spatial Ground Truth and Lane-Line Segmentation}
\label{sec:method:gt}

For each extracted frame \(I_t\), the GPS Map Camera application provides synchronized geo-temporal metadata, including latitude and longitude \(\mathbf{g}_t=[\phi_t,\lambda_t]^\top\) (and capture time, when available). These coordinates are used to query external map data and to associate each frame with a lane-count label.

\paragraph{\textbf{Lane-count ground truth from OpenStreetMap}}

We use the OpenStreetMap (OSM) Overpass API to retrieve lane-count information near each frame’s GPS location. For every coordinate \((\phi_t,\lambda_t)\), we query the nearest OSM road segment within a \textbf{15 m} radius and extract the number of lanes from OSM tags. If an explicit \texttt{lanes} tag is present, it is used directly; otherwise, we estimate the total as \(\texttt{lanes:forward}+\texttt{lanes:backward}\) when available. If no lane information is returned (or the query fails after retries), we mark the lane count as \texttt{N/A}. This yields a per-frame tuple
\[
\mathcal{D}=\{(I_t,\mathbf{g}_t,y_t)\}, \quad y_t \in \{1,2,4\}\ \text{or}\ \texttt{N/A}.
\]
To reduce noise from intermittent API failures, we discard rows with invalid coordinates or missing lane tags when constructing the labeled benchmark split.

\paragraph{\textbf{Lane-line segmentation}}

\begin{figure*}[t]
    \centering
    \begin{minipage}{0.32\textwidth}
        \includegraphics[width=\linewidth]{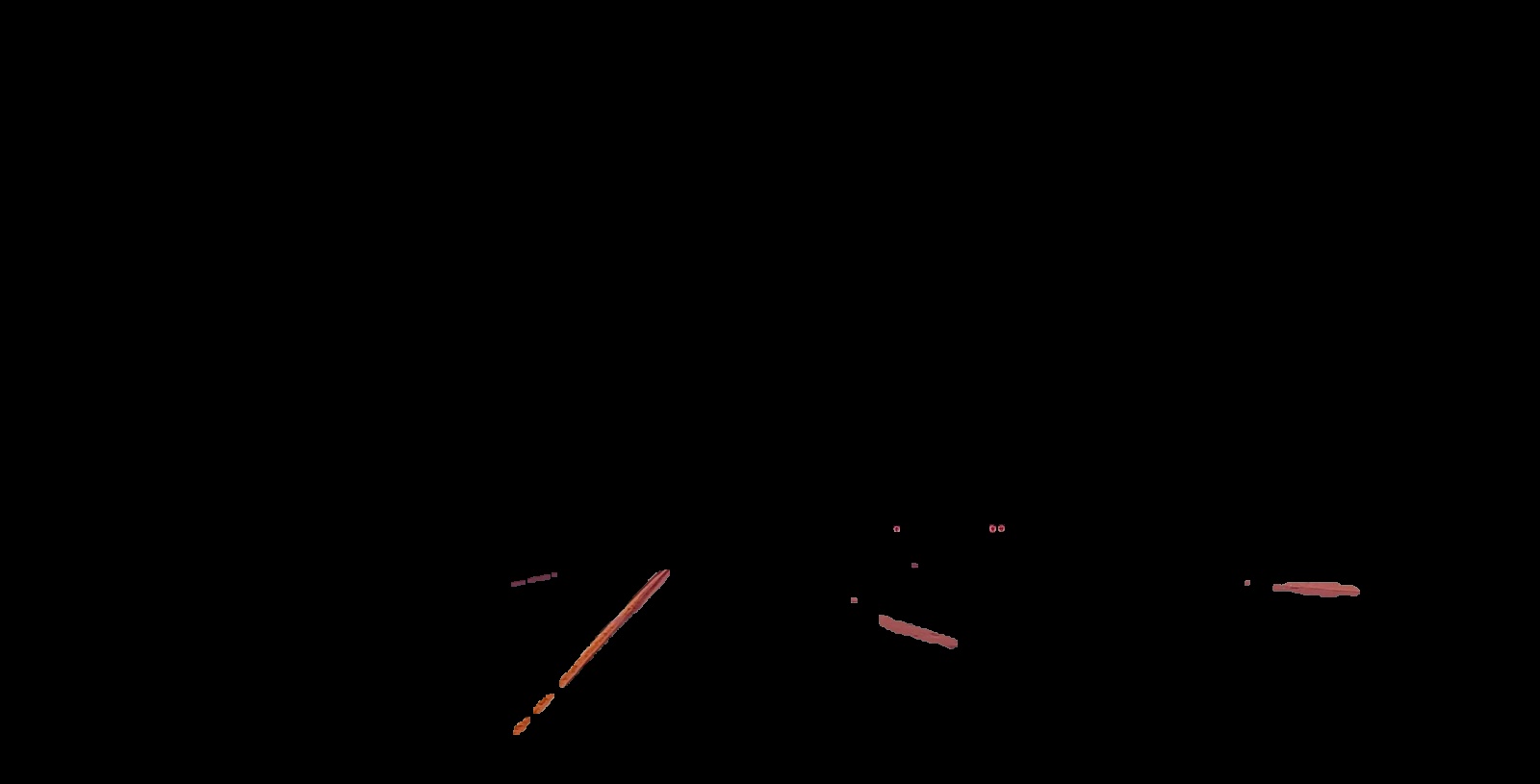}
        \centering {\small (a) 1 lane}
    \end{minipage}
    \hfill
    \begin{minipage}{0.32\textwidth}
        \includegraphics[width=\linewidth]{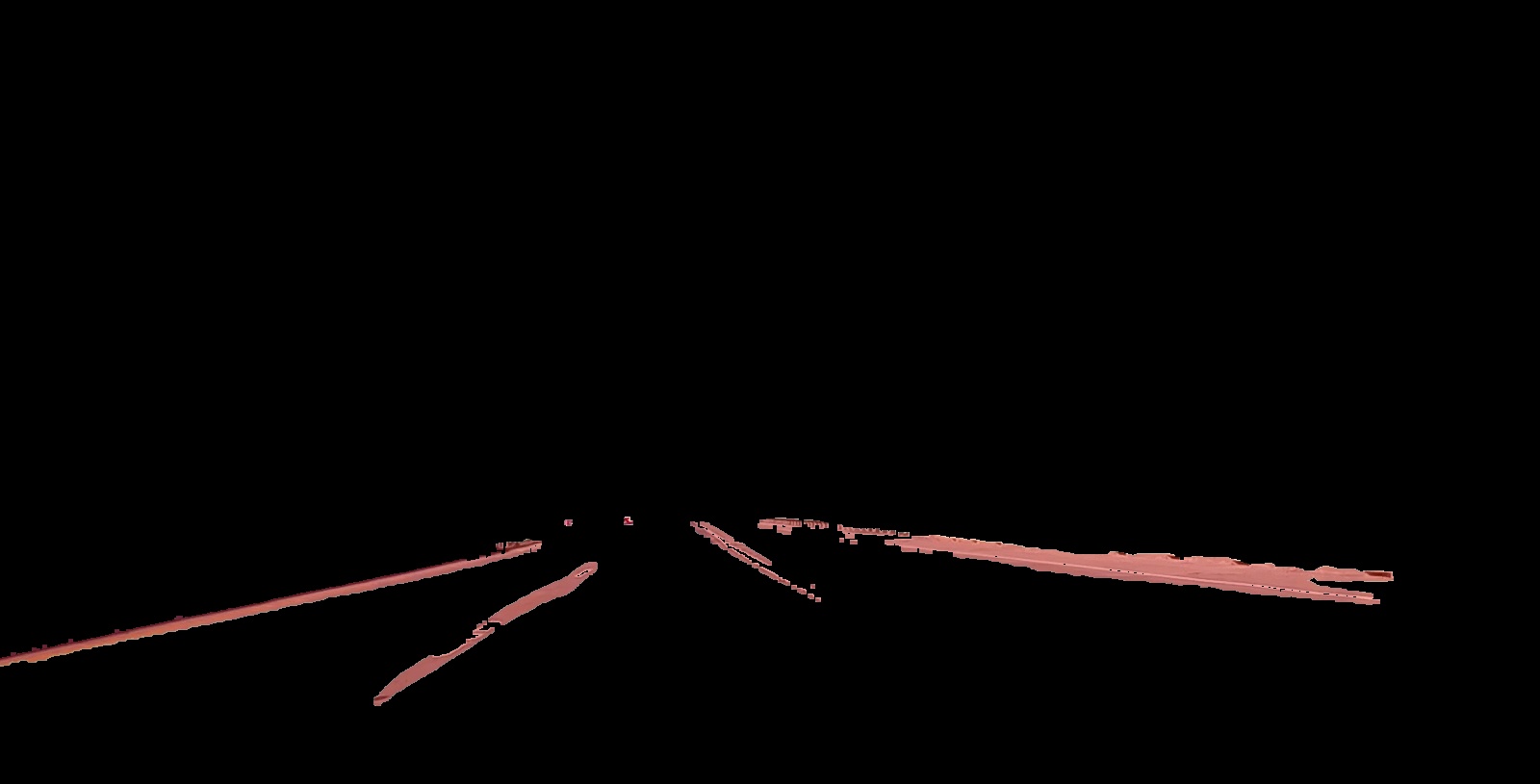}
        \centering {\small (b) 2 lanes}
    \end{minipage}
    \hfill
    \begin{minipage}{0.32\textwidth}
        \includegraphics[width=\linewidth]{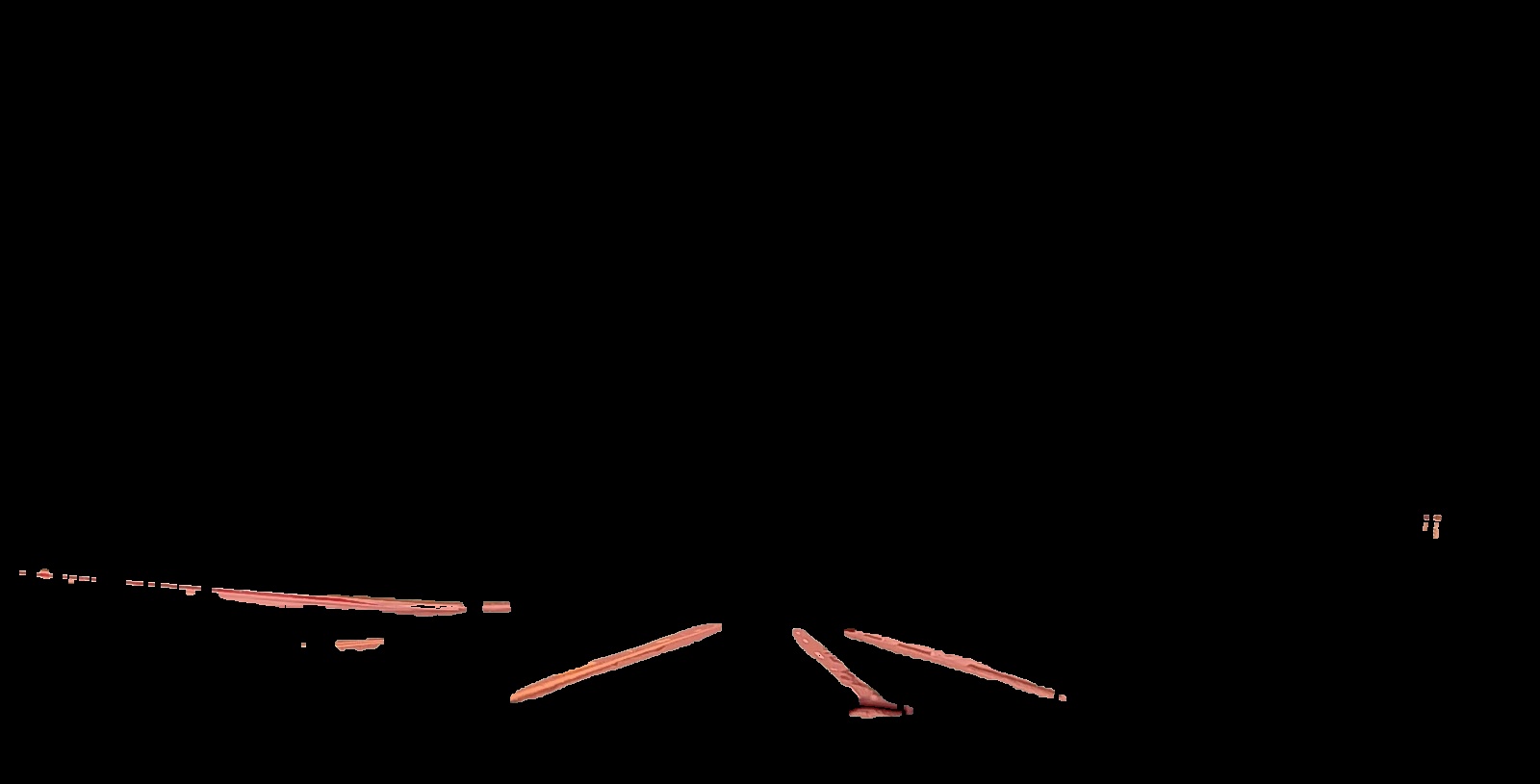}
        \centering {\small (c) 4 lanes}
    \end{minipage}

\caption{
Examples of YOLOP-segmented lane-line images, covering one- and two-lane road scenarios. 
}
    \label{fig:lane_seg_examples}
\end{figure*}

To isolate lane-marking pixels, we apply an off-the-shelf pretrained segmentation model called \textit{YOLOP} from Dong Wu et al. \cite{wu2022yolop}. Although YOLOP predicts both drivable-area and lane-line masks, we retain only the lane-line segmentation output for metric computation. The segmentation is not perfect (e.g., occasional leakage into road-surface regions), but it provides a consistent, automated lane-mask estimate across the data set. Figure~\ref{fig:lane_seg_examples} presents examples of segmentation output.

\paragraph{\textbf{Matched vs.\ Unmatched splits}}

The mapping API occasionally returns no lane-count data for certain coordinates. Therefore, we create two frame sets: (i) \textbf{Matched}, containing frames with valid OSM lane-count labels, and (ii) \textbf{Unmatched}, containing frames where the lane count is \texttt{N/A}. {In our current release, the total usable labeled images in \textbf{Matched} is \textbf{3,339}, with label counts \(\{1{:}196,\ 2{:}2872,\ 4{:}271\}\) (no class 3 observed). This subset serves as weak map-derived labels rather than manually verified annotations of the visible lane markings.}

\subsection{Problem Formulation}
\label{sec:method:formulation}

We introduce \textbf{three complementary lane quality metrics}, designed to capture correctness, visual clarity, and structural continuity of lane markings from dashcam imagery. These metrics fall into two categories:

\begin{itemize}
    \item \textbf{Correctness-based metric}: A lane-count prediction-based model that measures semantic agreement between the predicted lane count from the segmented image and ground-truth lane counts from the mapping API.
    
    \item {\textbf{No-reference Proxy metrics}}: Two image-driven regression metrics that quantify visual quality directly from segmented lane-line images, without requiring discrete labels.
    
\end{itemize}

All methods operate on the segmented lane-line image derived from each frame.

\subsubsection{Lane Count Prediction and Correctness}
\label{sec:method:lane_reg}

Given an input frame \( I_i \), we predict a \emph{continuous} lane count
\(\hat{y}_i \in \mathbb{R}\) to approximate the ground-truth lane count
\(y_i \in \{1,2,4\}\):
\begin{equation}
\hat{y}_i = f_{\theta}(I_i),
\end{equation}
where \( f_{\theta}(\cdot) \) is a regression model trained using a mean-squared error (MSE) loss.  
We formulate lane counting as regression rather than classification to preserve ordinal structure, handle class imbalance, and allow uncertainty-aware predictions. Instead of discrete accuracy, we measure semantic correctness using the distance between the predicted lane count value and the {ground truth} lane count from the mapping API. This formulation naturally yields a confidence score, defined later as the \textbf{Regression Lane Confidence Score (RLCS)}

\subsubsection{Lane Quality Scoring}
\label{sec:method:quality}

Beyond correctness, we quantify lane \emph{quality} using two transparent {no-reference, segmentation-derived lane-quality metrics}: \textbf{Contrast Stability Index (CSI)} and \textbf{Line Continuity Ratio (LCR)}.
{Each proposed metric is computed from an explicit formula, operates on the segmented lane-line
image derived from each frame, and is associated with a visually meaningful lane-marking property: CSI reflects contrast and visibility, LCR reflects geometric continuity, and RLCS reflects agreement between image-based lane-count prediction and map-derived lane-count information. Thus, the metrics are intended to be more transparent by design than black-box quality scores.}

\subsection{Contrast Stability Index (CSI)}
\label{sec:method:csi}

The Contrast Stability Index (CSI) measures the consistency of lane–background contrast by analyzing high-frequency image responses.  
Given a grayscale image \( G_i \) with segmented lanes, we compute its Laplacian:
\begin{equation}
L_i = \nabla^2 G_i,
\end{equation}
and define CSI as the standard deviation of the Laplacian response:
\begin{equation}
\mathrm{CSI}(I_i) = \sigma(L_i),
\end{equation}
where \( \sigma(\cdot) \) denotes standard deviation.

An optional normalized variant is:
\begin{equation}
\mathrm{CSI}_{\text{norm}}(I_i) = \frac{\sigma(L_i)}{\mu(G_i)+\epsilon},
\end{equation}
with mean intensity \( \mu(\cdot) \) and small constant \( \epsilon \).

\textbf{Interpretation:} Higher CSI indicates stronger and more consistent lane contrast, while lower values indicate faded paint, shadows, or poor illumination.

\subsection{Line Continuity Ratio (LCR)}
\label{sec:method:lcr}

The Line Continuity Ratio (LCR) quantifies the geometric completeness of lane markings.   We first compute an edge map \(E_i\) using the Canny operator, followed by probabilistic Hough line detection to obtain a set of line segments \(\mathcal{H}_i\).

Let each segment \(j \in \mathcal{H}_i\) have endpoints \(p_1^j\) and \(p_2^j\). The total detected lane length is:
\begin{equation}
\ell_i = \sum_{j \in \mathcal{H}_i} \|p_1^j - p_2^j\|_2,
\end{equation}
and LCR is defined as:
\begin{equation}
\mathrm{LCR}(I_i) = \frac{\ell_i}{|E_i| + \epsilon},
\end{equation}
where \(|E_i|\) is the number of edge pixels.

\textbf{Interpretation:} Higher LCR corresponds to longer, smoother, and less fragmented lane markings. A lower LCR would suggest fragmented or poor-quality lane markings. However, improving the lane segmentation algorithm and normalizing for variation in weather could make this even more reliable.

\subsection{Regression Lane Confidence Score (RLCS)}
\label{sec:method:rlcs}

{The Regression Lane Confidence Score measures agreement between the predicted lane count and the OSM-derived lane-count label. For an OSM-matched frame $I_i$, let $y_i$ denote the lane count obtained from OSM and let  $\hat{y}_i=f_\theta(I_i)$ denote the continuous lane-count prediction. We compute prediction error as: $d_i = |\hat{y}_i - y_i|$. The RLCS is then defined as:
\begin{equation}
\mathrm{RLCS}(I_i) = \exp\left(-\frac{d_i^2}{\tau^2}\right),
\end{equation}
where \(\tau>0\) controls  controls tolerance to regression error.}

\textbf{Interpretation:} {A prediction close to the OSM-derived lane count receives a high score, while a prediction close to an incorrect lane-count category receives a low score. For example, predicting four lanes for a two-lane road produces a large error and therefore a low RLCS. RLCS is only defined for frames with valid OSM-derived lane-count labels. RLCS thus can be considered as a correctness-aware semantic quality metric, complementary to CSI and LCR.}

\section{Results}
\label{sec:results}

We evaluate the proposed lane quality framework along three dimensions:
(1) lane-count prediction accuracy and semantic confidence,
(2) percentile-based prediction of contrast stability (CSI),
and (3) percentile-based prediction of line continuity (LCR).
All experiments are conducted on segmented lane images obtained using a pretrained YOLOP model.
Deep models are trained on an NVIDIA RTX 3080 Ti GPU using ImageNet-pretrained MobileNetV2 backbones.

\begin{figure}[t]
    \centering
    \includegraphics[width=\linewidth]{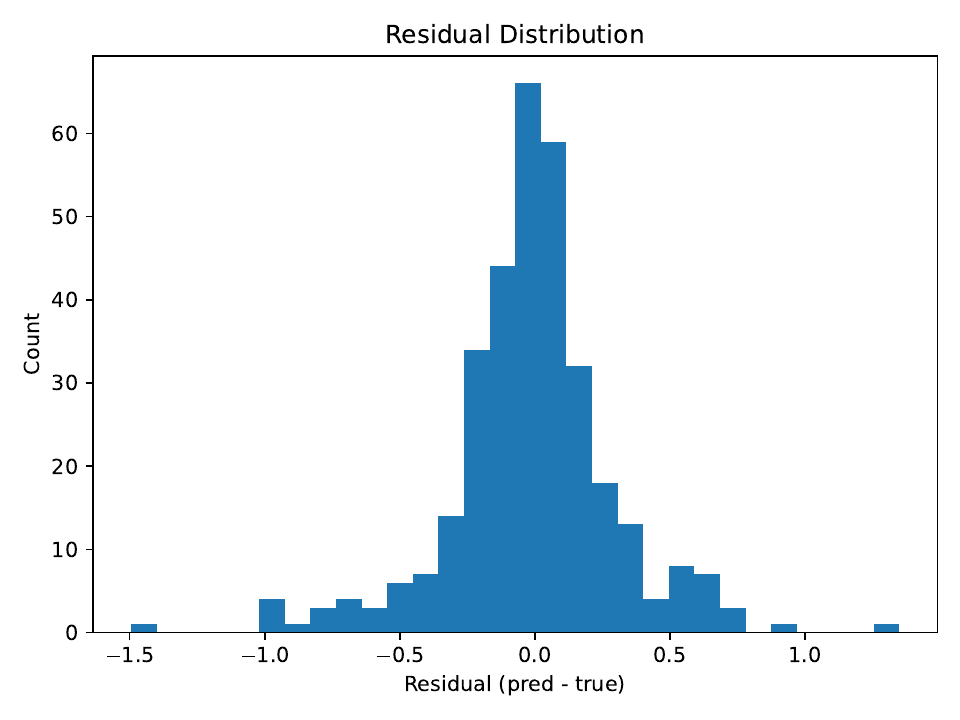}
    \caption{{Residual distribution for lane-count regression.
    Most predictions cluster near zero error, with tails caused by segmentation noise, occlusions, and mapping failures.}}
    \label{fig:residual}
\end{figure}

\begin{figure}[t]
    \centering
    \includegraphics[width=\linewidth]{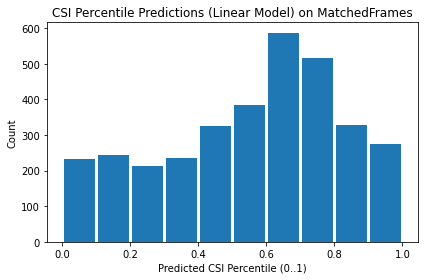}
    \caption{Distribution of predicted CSI percentiles on the MatchedFrames set. Higher percentiles indicate stronger and more stable lane–background contrast, while lower values correspond to faded markings, glare, or poor illumination.}
    \label{fig:csi_dist}
\end{figure}

\begin{figure}[htbp]
    \centering
    \includegraphics[width=\linewidth]{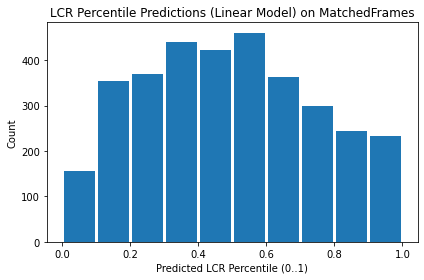}
    \caption{Distribution of predicted LCR percentiles on the MatchedFrames set.
    Higher percentiles correspond to long, continuous lane markings, whereas lower values indicate fragmented or discontinuous lane structures.}
    \label{fig:lcr_dist}
\end{figure}

\subsection{Lane Count Prediction and Correctness-Based Quality}
\label{sec:results:lane_count}

We first evaluate the lane-count regression model trained on the \textit{Matched} subset, where OpenStreetMap (OSM) provides reliable lane-count ground truth.
The model predicts a continuous lane count, which is later used to derive the
\emph{Regression Lane Confidence Score (RLCS)}. Due to class imbalance in the lane count (the majority of frames contain two lanes) we have used class weights to improve the model training. After 20 epochs of training, the regression model achieves:
\begin{itemize}
    \item \textbf{MSE}: 0.096
    \item {\textbf{R$^{2}$}: 0.751}
\end{itemize}

As illustrated in Figure~\ref{fig:residual}, the model is still affected by occasional segmentation noise, but it provides a meaningful continuous estimate that supports downstream confidence scoring.

\begin{figure*}[t]
\centering

\begin{minipage}{0.19\linewidth}
\includegraphics[width=\linewidth]{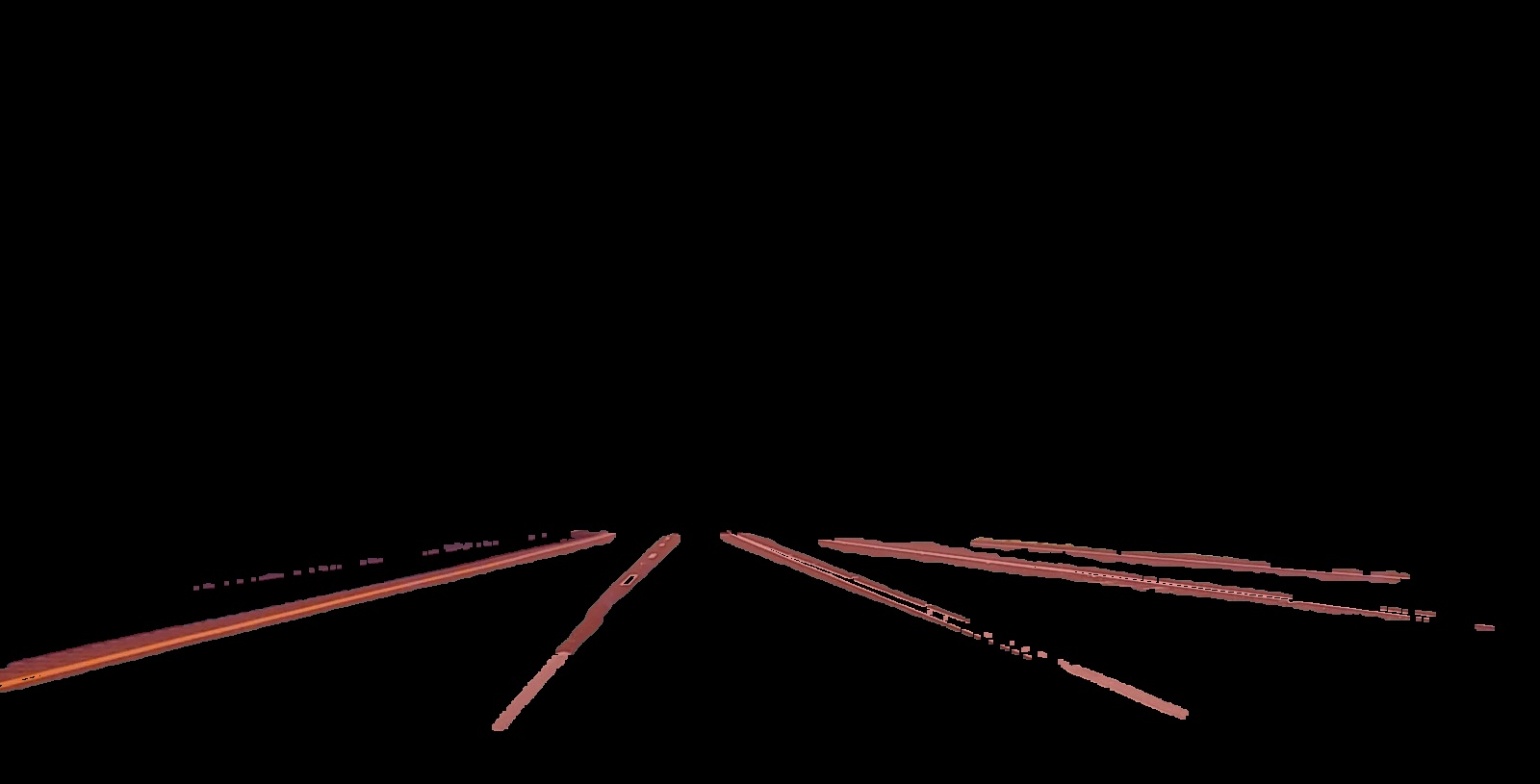}
\end{minipage}
\hfill
\begin{minipage}{0.19\linewidth}
\includegraphics[width=\linewidth]{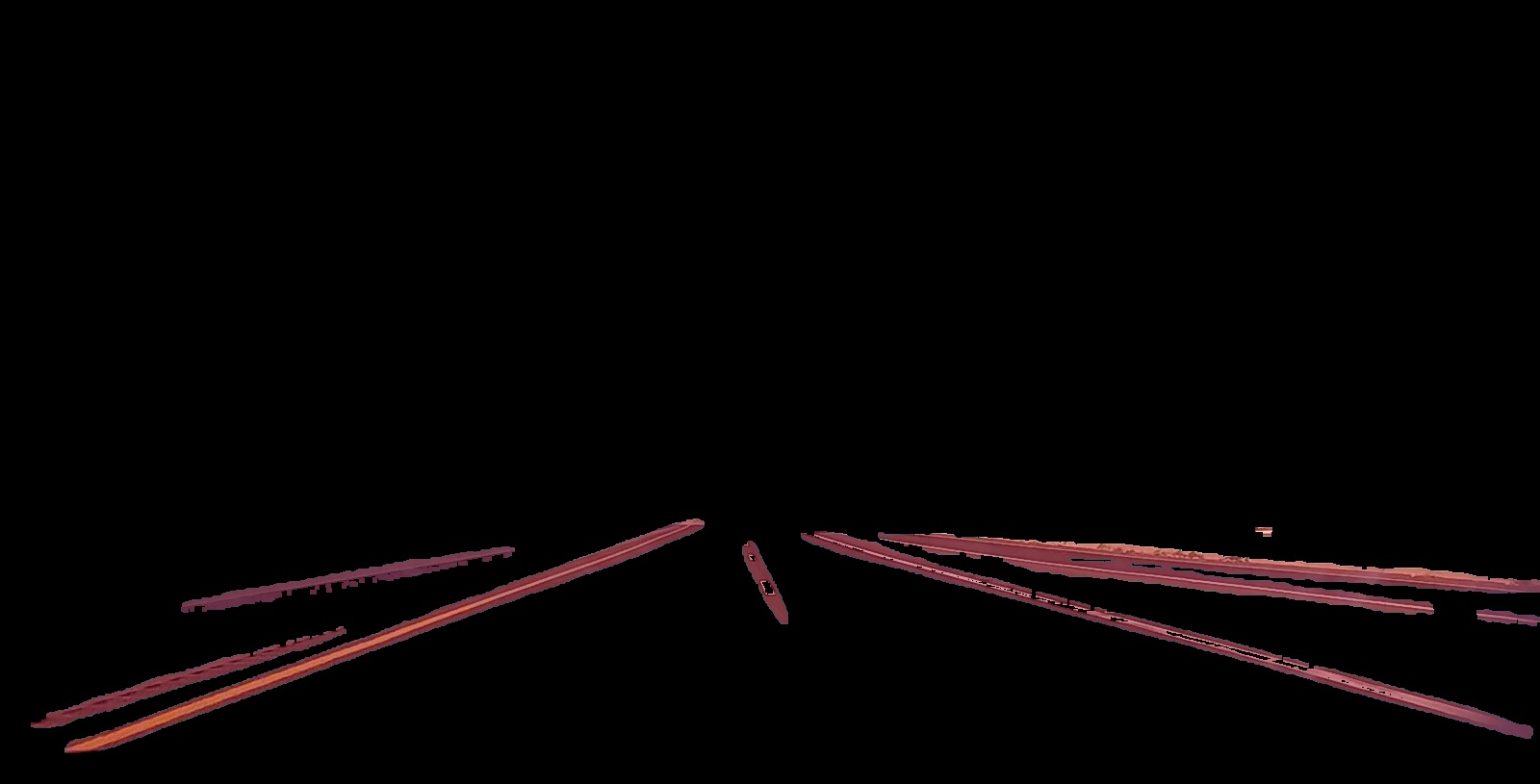}
\end{minipage}
\hfill
\begin{minipage}{0.19\linewidth}
\includegraphics[width=\linewidth]{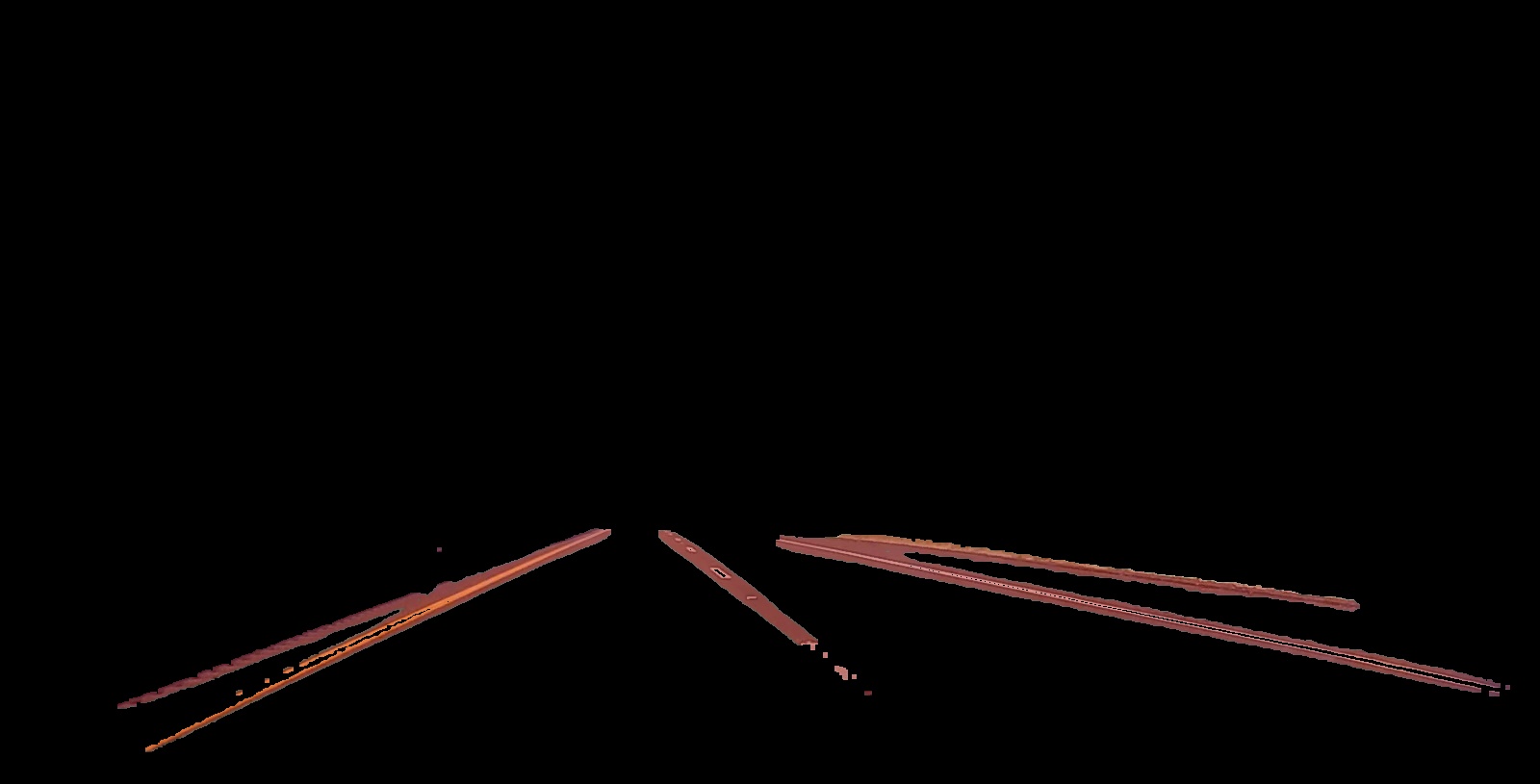}
\end{minipage}
\hfill
\begin{minipage}{0.19\linewidth}
\includegraphics[width=\linewidth]{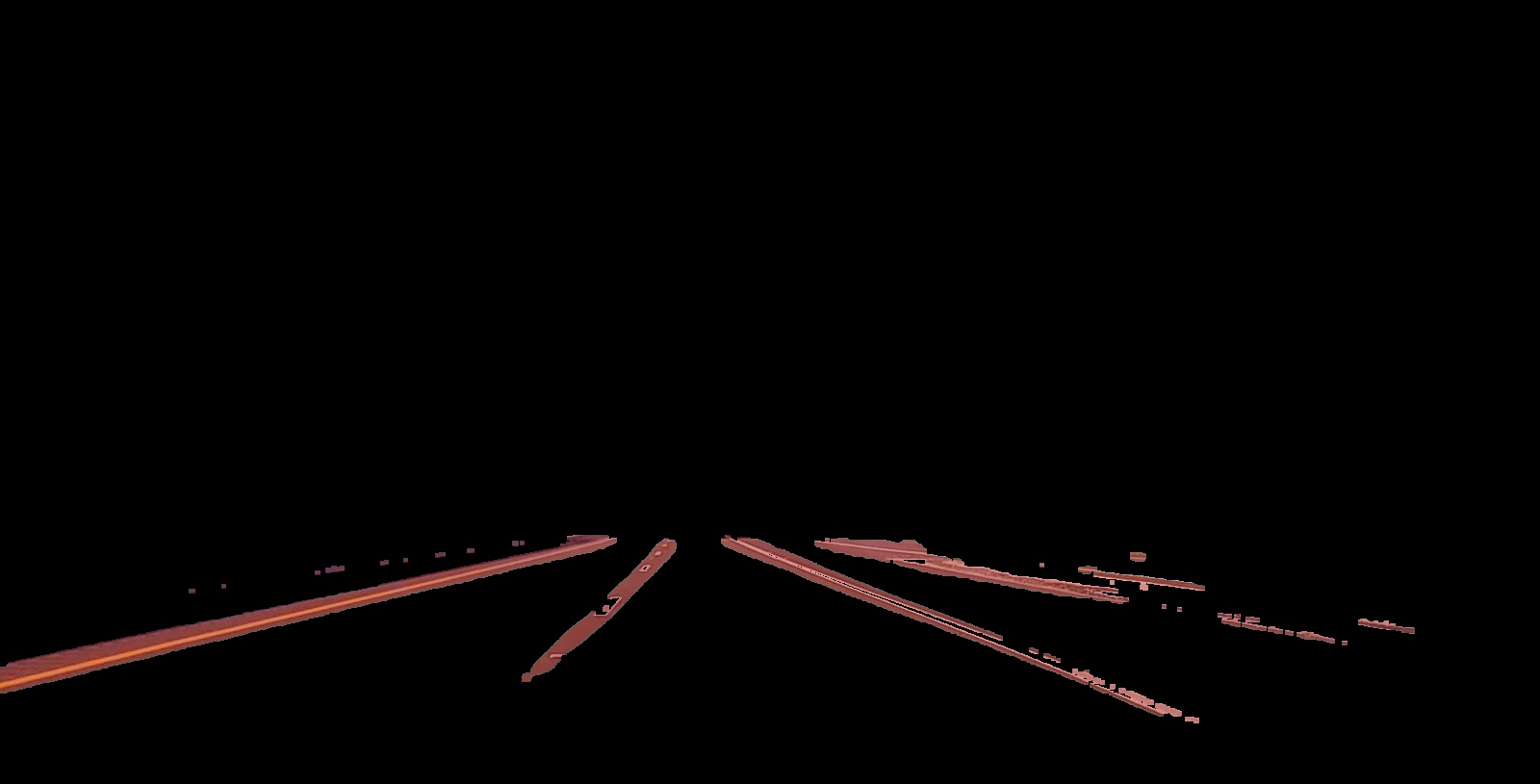}
\end{minipage}
\hfill
\begin{minipage}{0.19\linewidth}
\includegraphics[width=\linewidth]{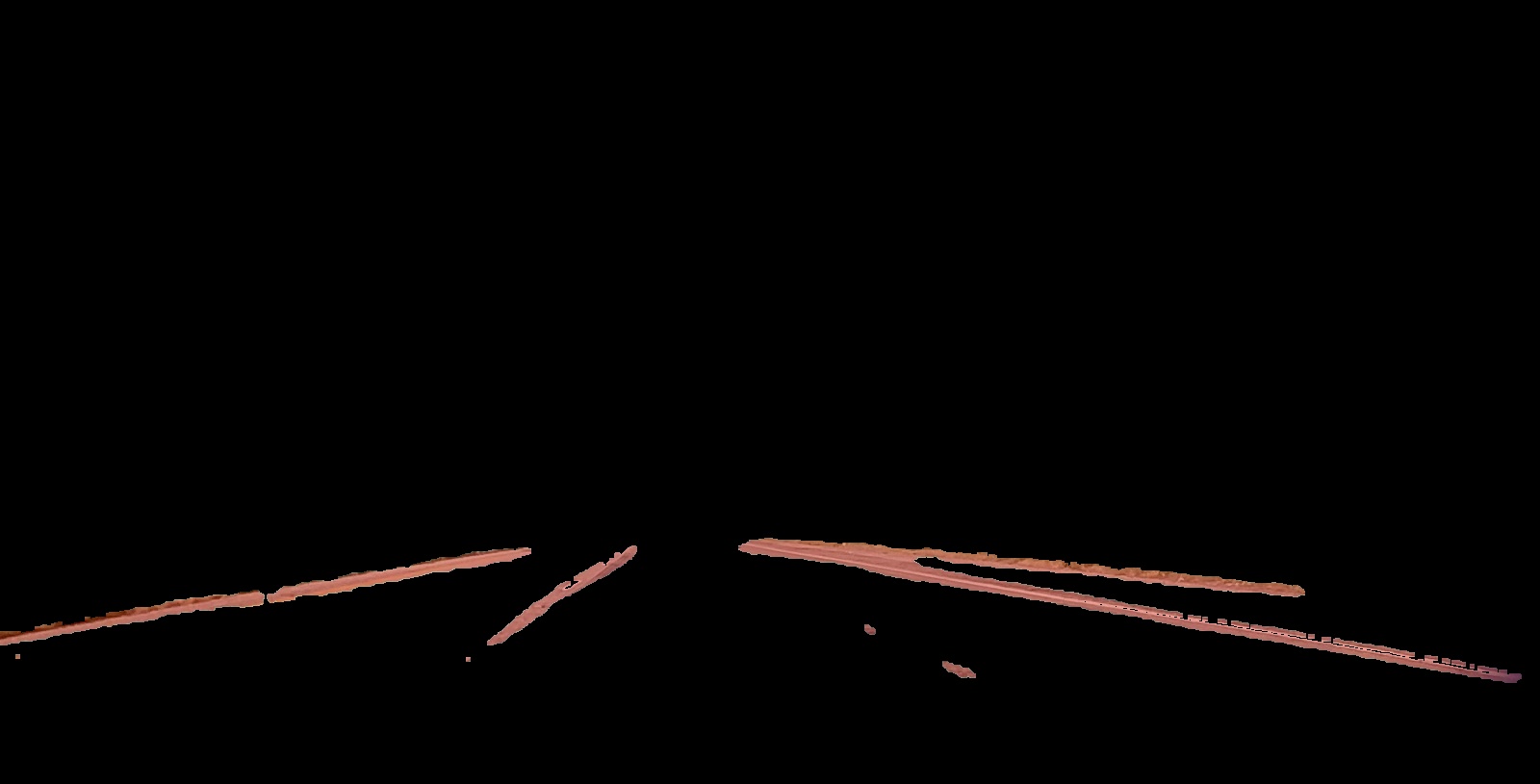}
\end{minipage}

\vspace{.5 cm }

\begin{minipage}{0.19\linewidth}
\includegraphics[width=\linewidth]{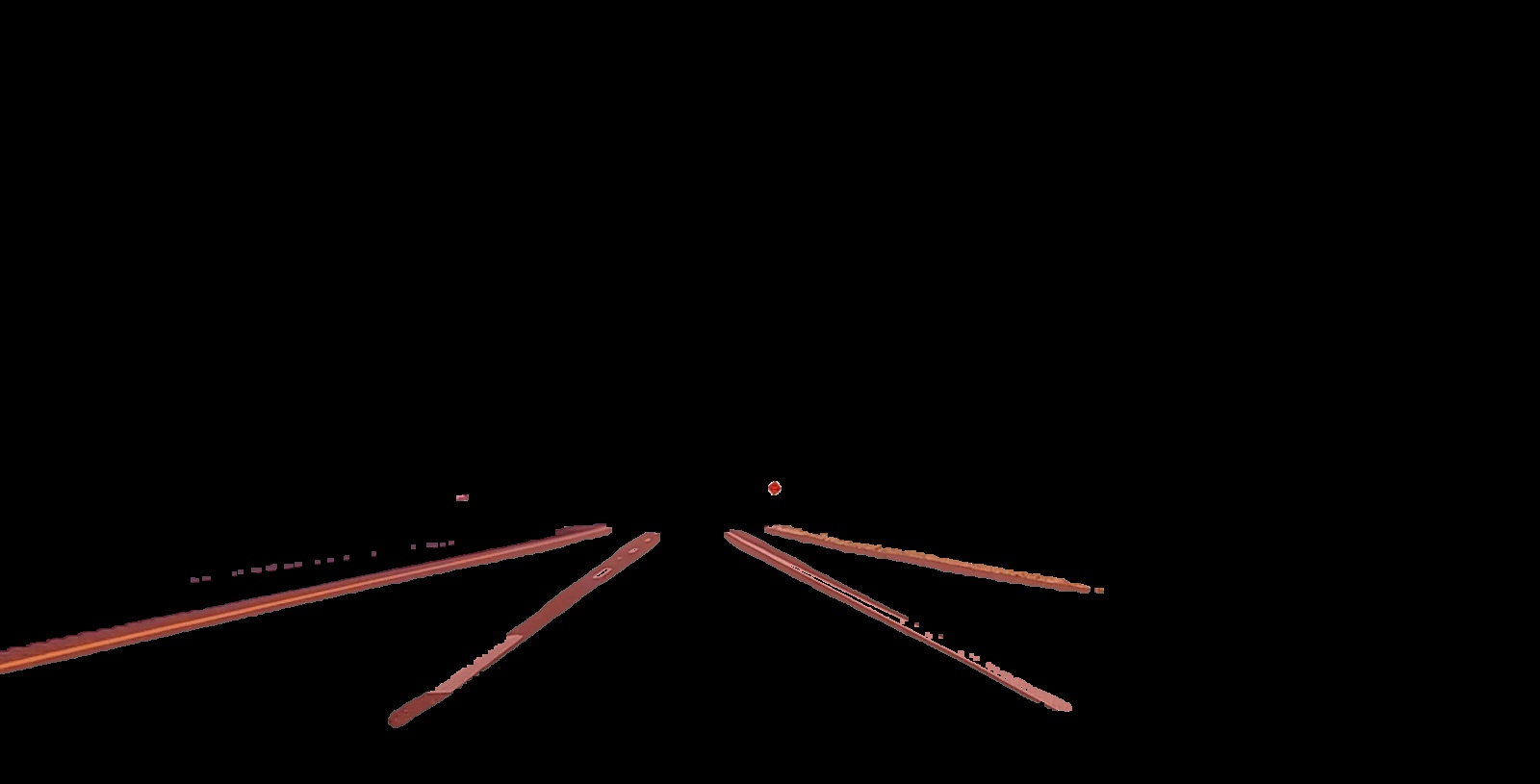}
\end{minipage}
\hfill
\begin{minipage}{0.19\linewidth}
\includegraphics[width=\linewidth]{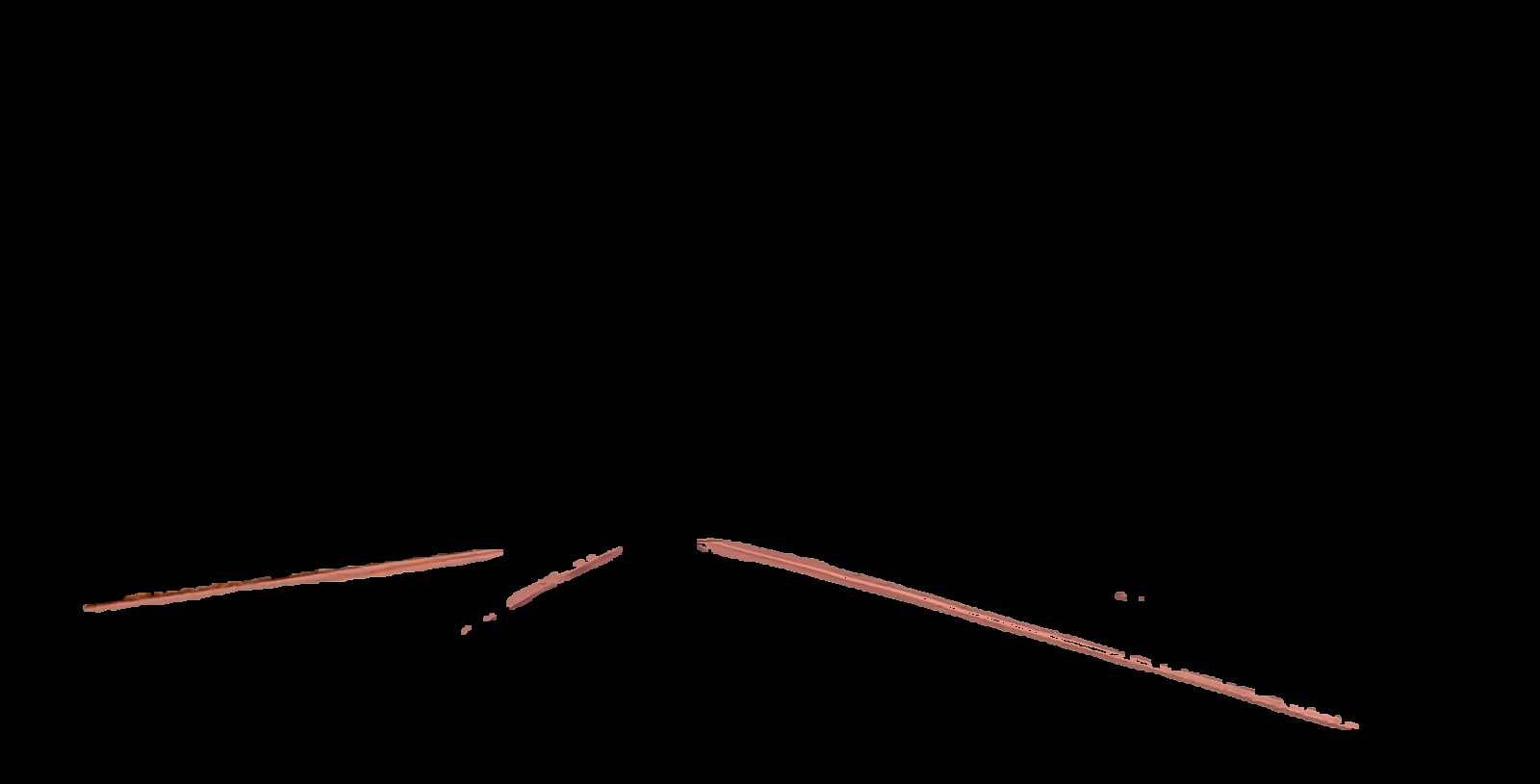}
\end{minipage}
\hfill
\begin{minipage}{0.19\linewidth}
\includegraphics[width=\linewidth]{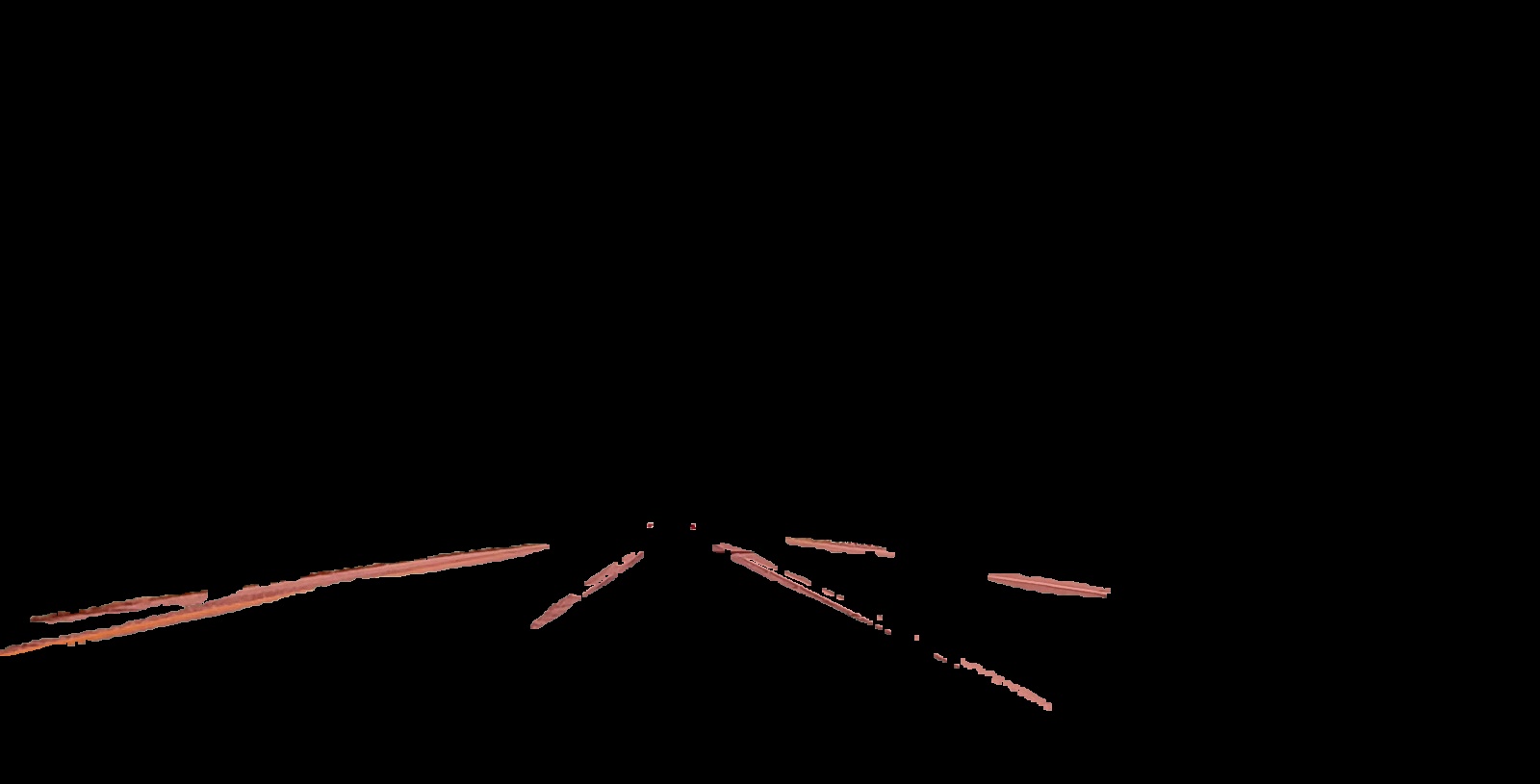}
\end{minipage}
\hfill
\begin{minipage}{0.19\linewidth}
\includegraphics[width=\linewidth]{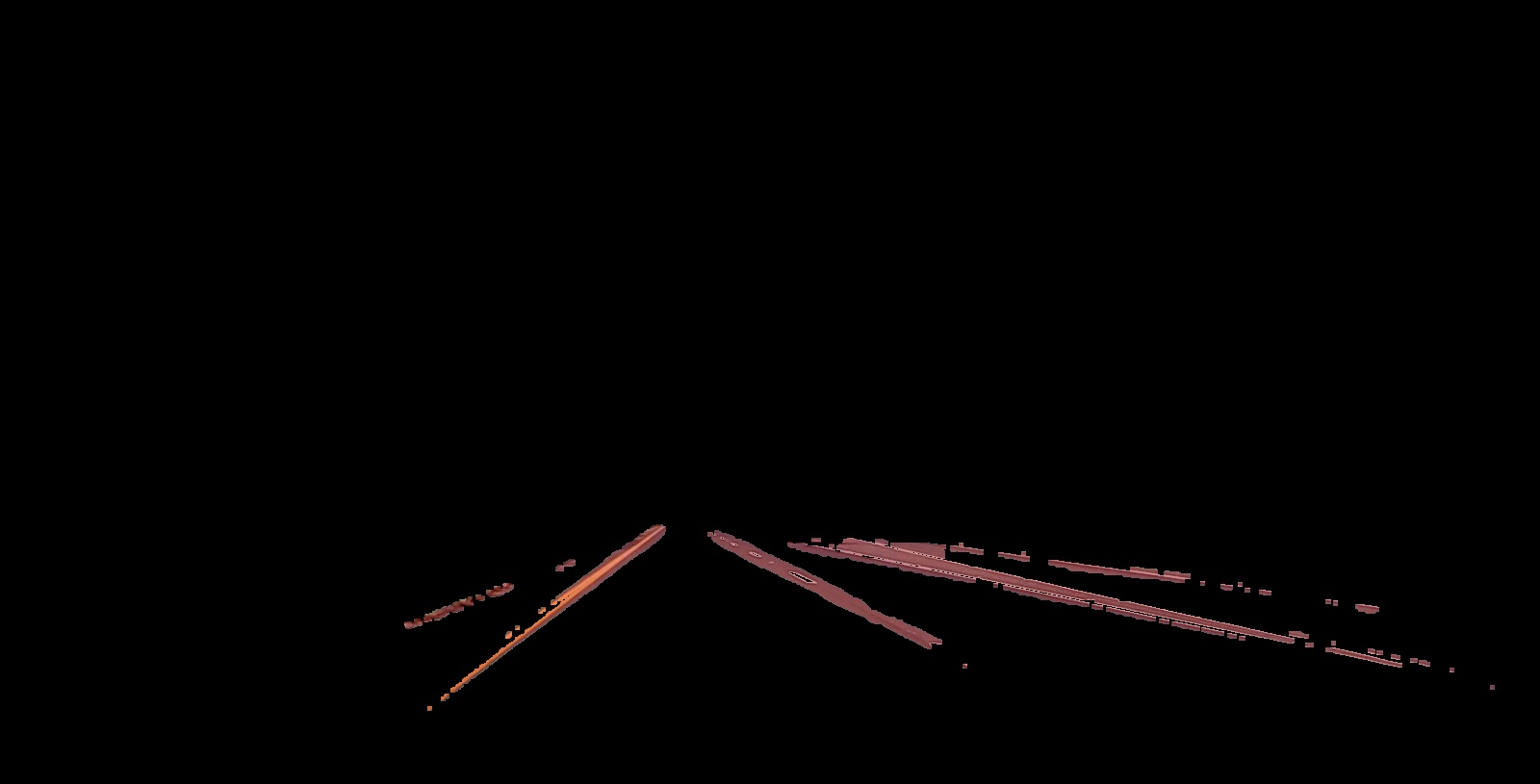}
\end{minipage}
\hfill
\begin{minipage}{0.19\linewidth}
\includegraphics[width=\linewidth]{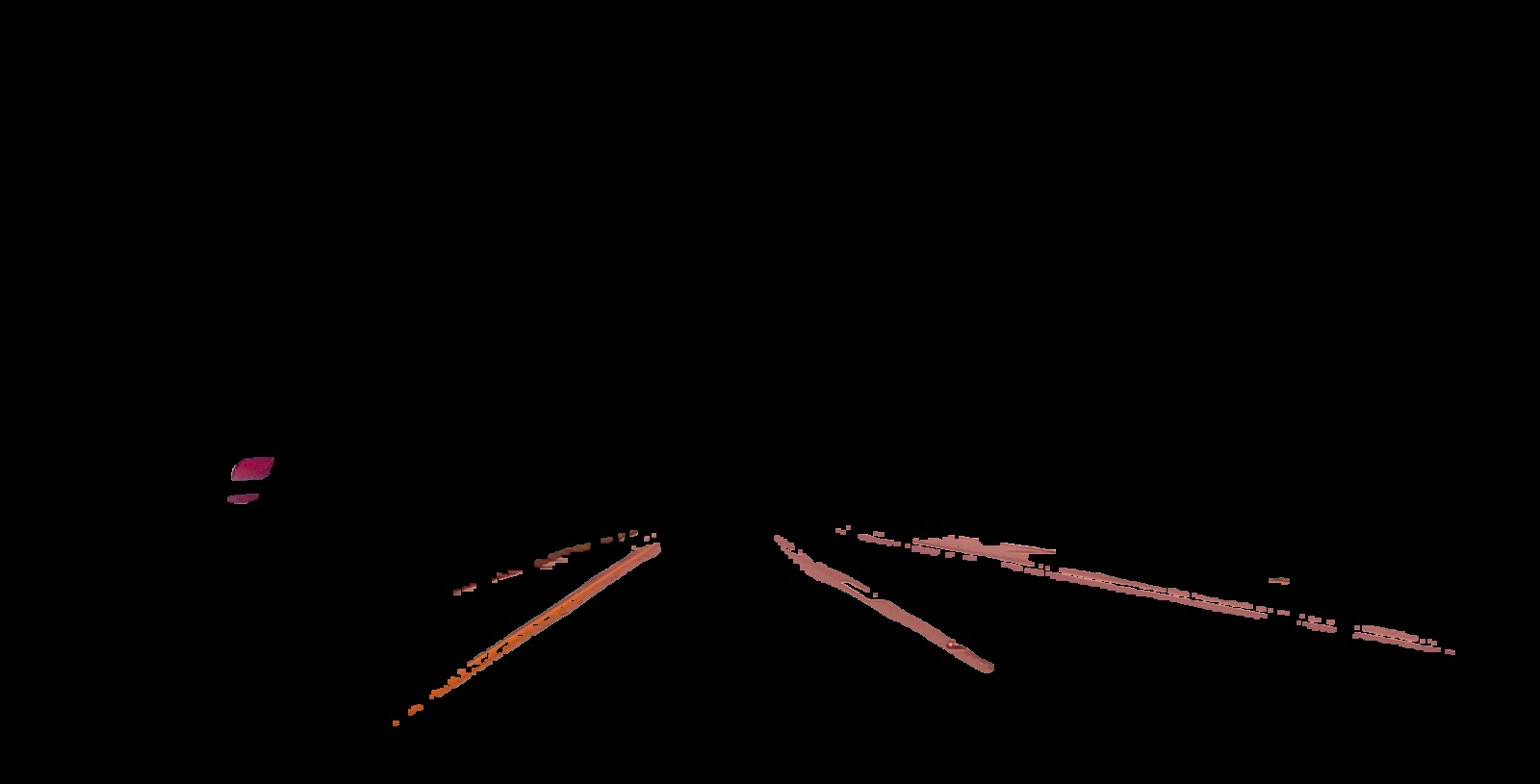}
\end{minipage}

\caption{Ten example images manually ranked from the highest lane quality (left) to the lowest lane quality (right). These images are used to validate the ranking consistency of the proposed CSI and LCR metrics against human assessment.}
\label{fig:ranked_images}
\end{figure*}

\subsection{Percentile-Based CSI and LCR Prediction}
\label{sec:results:percentile}

For CSI and LCR, no absolute ground truth exists.
Instead, we train regression models to predict \emph{percentile-normalized} scores in $[0,1]$,
learned from large collections of unlabeled (\textit{Unmatched}) frames.
These models are evaluated qualitatively through score distributions on the \textit{Matched} set.

Figure~\ref{fig:csi_dist} shows the predicted CSI percentile distribution.
Most frames concentrate in the mid-to-high percentile range, suggesting that
the model captures meaningful contrast variation across different road conditions.
Lower percentiles correspond to faded paint, glare, or poor illumination,
while higher percentiles reflect strong lane–background contrast.

Figure~\ref{fig:lcr_dist} presents the predicted LCR percentile distribution.
Compared to CSI, LCR exhibits a broader spread, reflecting greater structural variability
in lane continuity due to occlusions, road curvature, and segmentation noise.
Higher LCR percentiles correspond to long, uninterrupted lane markings, while lower percentiles indicate fragmented or missing lanes.

The CSI-percentile regression model achieves strong performance with MAE of 0.039, MSE of 0.0021, and $R^2$ of 0.97. {These indicate that the baseline model can well approximate the proposed formula-derived CSI score.} The LCR-percentile model has slightly higher error, with MAE of 0.093 and MSE of 0.013 ($R^2$ = 0.84). This is probably due to greater structural variability and difficulty in modeling lane continuity. {The separate pilot human-ranking experiment discussed in Sec. \ref{sec:validation} provides complementary evidence that CSI and LCR align with perceived lane-marking quality.}

\subsection{CSI and LCR Prediction Validation Using Human Ground Truth}
\label{sec:validation}

Although CSI and LCR are designed as transparent lane-quality metrics, it is important to evaluate whether they can predict the perception of lane quality. To provide a ground-truth validation experiment, we selected \textbf{50 images} from the test set and ranked them according to the quality of their lane markings. The images were ordered from the best lane markings to the poor-quality lane markings.

To evaluate how well the predicted CSI and LCR percentile scores preserve this ordering, we employ two well-known ranking-based evaluation metrics.

\begin{itemize}
    \item \textbf{Spearman Rank Correlation ($\rho$):} Measures the strength of the monotonic relationship between the predicted ranking and the human ranking \cite{zar2005spearman}.
    
    \item \textbf{Kendall's Tau ($\tau$):} Measures the ordinal agreement between two rankings by counting the number of concordant and discordant image pairs\cite{noether1981kendall}.
\end{itemize}

We further compare the proposed metrics against several commonly used image quality measures.

\begin{itemize}

\item \textbf{Peak Signal-to-Noise Ratio (PSNR):} Measures quality through reconstruction fidelity using pixel-wise intensity differences, where higher values indicate greater similarity.

\item \textbf{Structural Similarity Index Measure (SSIM):} Evaluates perceptual image similarity based on luminance, contrast, and structural information\cite{brunet2011mathematical}.

\item \textbf{Blind/Referenceless Image Spatial Quality Evaluator (BRISQUE):} A no-reference image quality metric that estimates perceptual quality using natural scene statistics\cite{mittal2011blind}.

\item \textbf{Natural Image Quality Evaluator (NIQE):} A completely unsupervised no-reference image quality metric based on deviations from natural image statistics\cite{gunawan2025optimizing}.

\item \textbf{Learned Perceptual Image Patch Similarity (LPIPS):} Measures perceptual image similarity using deep feature representations extracted from pretrained neural networks\cite{zhang2018unreasonable}.
    
\end{itemize}

\begin{table}[t]
\centering
\caption{Ranking agreement with manually assigned lane-quality rankings. Spearman's $\rho$ and Kendall's $\tau$ measure agreement with the human ranking. Positive correlations indicate direct agreement, while negative correlations are expected for BRISQUE, NIQE, and LPIPS because lower scores indicate better image quality. The proposed \textbf{LCR} achieves the strongest agreement with human perception, followed by \textbf{CSI}, outperforming all conventional image quality metrics.}
\label{tab:ranking_validation}
\begin{tabular}{lcc}
\toprule
\textbf{Method} & \textbf{Spearman's $\rho$} & \textbf{Kendall's $\tau$} \\
\midrule
LCR      & \textbf{0.7316} & \textbf{0.5472} \\
CSI      & 0.6112 & 0.4400 \\
\hline
NIQE     & -0.5699 & -0.3959 \\
LPIPS    & -0.5251 & -0.3878 \\
SSIM     & -0.4504 & -0.3469 \\
PSNR     & -0.2526 & -0.1804 \\
BRISQUE  & -0.2279 & -0.1535 \\

\bottomrule
\end{tabular}
\end{table}

Figure~\ref{fig:ranked_images} shows an example of manually ranked images used for evaluation. Table~\ref{tab:ranking_validation} demonstrates that the proposed lane-quality metrics are in stronger agreement with human classification than conventional image quality measures. Both CSI and LCR obtain Spearman's rank correlation and Kendall's Tau higher than those of SSIM and PSNR, indicating that the proposed metrics better capture the perceptual quality of lane markings. 

\subsection{Limitations and Future Work}

{The current study has several limitations. First, the dataset is geographically limited to South Texas and was collected primarily during daytime under clear or moderately cloudy conditions. Therefore, the current results may not generalize to night driving, rain, snow, fog, or regions with different road-marking standards. Second, OSM lane-count labels are weak map-derived labels rather than manually verified visual annotations, and only 3,339 frames in the current release are matched to valid lane-count information. Third, the human validation set contains only 50 manually ranked images and should be interpreted as pilot evidence rather than comprehensive expert validation. Fourth, CSI and LCR depend on the quality of the lane-line segmentation model; errors from YOLOP can propagate into the proposed quality metrics, especially for faded or occluded lane markings. Fifth, although the proposed metrics are explicit and physically motivated, their utility has not been validated through a formal study with transportation engineers or road-maintenance experts. Future work should expand the dataset across regions, lighting, weather, and road types; add expert lane-quality annotations and validation to proxy metrics; evaluate robustness to segmentation errors; and investigate the utility of the existing autonomous-driving datasets for road lane quality assessment.}

\section{Conclusions}

In this work, we present the first open-source dataset, benchmark, and AI toolset for lane marking quality assessment using geo-tagged smartphone dashcam data. Assessing lane marking quality is a difficult problem because there is no human annotation, and the evaluation has to rely on comparison. To address this, we introduce simple and transparent metrics, CSI, LCR, and RLCS, that capture contrast, continuity, and semantic correctness without dense labels. The proposed models show strong results, with CSI prediction being highly accurate (MAE 0.039, $R^2$ 0.97) and LCR also reasonably reliable (MAE 0.093, $R^2$ 0.84), and we expect this to improve further with better lane segmentation since the current setup relies on an off-the-shelf model. {Overall, the open-source data, toolset, and baseline provide a foundational framework for this problem, facilitating future work on scalable lane quality monitoring.}



\bibliographystyle{ieeetr}
\bibliography{ref}

\end{document}